\documentclass[10pt]{article}
\usepackage[letterpaper,textwidth=6.0in,textheight=9in,centering]{geometry}
\usepackage{times}
\usepackage[round,authoryear]{natbib}
\usepackage[T1]{fontenc}
\usepackage[utf8]{inputenc}
\usepackage{microtype}
\usepackage{amsmath,amssymb}
\usepackage{booktabs}
\usepackage[table]{xcolor}
\usepackage{arydshln}
\usepackage{multirow}
\usepackage{graphicx}
\usepackage{wrapfig}
\usepackage{needspace}
\usepackage{enumitem}
\usepackage{algorithm}
\usepackage{algpseudocode}
\usepackage{placeins}
\usepackage{xurl}
\usepackage{listings}
\usepackage{hyperref}
\usepackage{cleveref}
\usepackage{tikz}
\usetikzlibrary{arrows.meta,positioning,fit,backgrounds}

\definecolor{inheritblue}{HTML}{176B87}
\definecolor{inheritgreen}{HTML}{2D7D46}
\definecolor{inheritred}{HTML}{A83A32}
\definecolor{lightgray}{HTML}{F2F4F5}
\hypersetup{
 colorlinks=true,
 linkcolor=inheritblue,
 citecolor=inheritgreen,
 urlcolor=inheritblue,
 pdftitle={Inherit-MAS: Test-Time Evolution of Multi-Agent Systems through Workflow and Execution Inheritance},
 pdfauthor={Songtao Wei}
}
\setlist{nosep,leftmargin=*}
\newcommand{\live}{\ensuremath{T_{\mathrm{live}}}}
\newcommand{\inherited}{\ensuremath{T_{\mathrm{inherited}}}}
\newcommand{\paddedhdashline}{%
 \noalign{\vskip\aboverulesep}%
 \hdashline
 \noalign{\vskip\belowrulesep}%
}

\title{Inherit-MAS: Test-Time Evolution of Multi-Agent Systems through Workflow and Execution Inheritance \\

}
\author{Songtao Wei$^\alpha$, Yi Li$^\alpha$, Zhichun Guo$^\beta$, and Bingzhe Li$^\alpha$\\
$^\alpha$University of Texas at Dallas\\
$^\beta$Independent researcher\\
\texttt{songtao.wei@utdallas.edu}\\
Code: \url{https://github.com/CrazyMint/Inherit-MAS}}
\date{}

\begin{document}
\maketitle

\begin{abstract}

Multi-agent systems (MAS) built from large language models coordinate specialized agents to tackle complex tasks, but effective workflows are difficult to design in advance. Test-time evolution refines workflows using execution feedback, yet broad revisions can disturb useful components, while re-executing unchanged requests can incur redundant computation. Inspired by the interplay of inheritance and selection in biological evolution, we introduce Inherit-MAS, which makes inheritance explicit at the workflow and execution levels. A meta-model first synthesizes a workflow of worker agents with declared roles, communication inputs, and tool permissions, and a separately prompted judge scores each executed candidate and diagnoses its deficiencies. In ordinary refinement rounds, \emph{workflow inheritance} starts from the latest completed candidate, may discard removable nodes judged unhelpful, and applies a validated edit to address the diagnosed deficiency. When the new candidate executes, \emph{execution inheritance} inherits eligible stored results only if the complete resolved request and execution context match, avoiding redundant model and tool calls. With GPT-4o-mini workers, Inherit-MAS achieves 55.4\% completion on WorkBench and 49.7\% joint F1 on HotpotQA FullWiki, outperforming EvoAgent, EvoMAS, and TacoMAS. With Qwen3-32B workers, it also exceeds these evolving-MAS baselines on both benchmarks. Compared with rerunning the same controller with execution inheritance disabled, execution inheritance reduces worker-token usage by 29.1\% on WorkBench and 34.6\% on HotpotQA, and total token usage by 5.3\% and 18.1\%.

\end{abstract}

\section{Introduction}
Large language model (LLM) agents are increasingly moving beyond single-turn question answering toward complex tasks that require planning, tool use, interaction with external environments~\citep{yao2023react}, and adaptation over extended periods of time~\citep{shinn2023reflexion}. One approach to managing these demands is to coordinate specialized language-model agents in a multi-agent system (MAS), using workflows for decomposition, tool use, communication, and verification. Rather than asking one LLM to maintain the entire reasoning and execution trajectory, a MAS can divide the task among agents with different responsibilities, execute independent subtasks in parallel, exchange intermediate results, and assign other agents to critique or verify them. Systems such as AutoGen~\citep{wu2024autogen} and MetaGPT~\citep{hong2024metagpt} instantiate these patterns through programmable conversations and role-based procedures, while automatic design methods search agent programs~\citep{hu2024adas}, executable workflows~\citep{zhang2024aflow}, prompts, and graph connections~\citep{zhuge2024gptswarm}. Across these approaches, performance depends on task decomposition, role and tool assignment, information flow, and verification. Because their interactions are difficult to anticipate, the initial execution may expose weaknesses in evidence collection, tool use, or coordination.


However, such diagnostic information becomes available only after the initial workflow has been synthesized and executed. A fixed workflow can respond to observations within its prescribed execution pattern, but it cannot revise that pattern to address weaknesses exposed by a particular task. Evolving-MAS methods aim to close this gap by revising or modifying an initial system rather than treating it as final, through evolutionary operations over an expert agent~\citep{yuan2024evoagent}, pools of promising configurations with accumulated experience~\citep{hu2026evomas}, or coupled fast and slow test-time loops~\citep{xu2026tacomas}. Yet deciding what to change is only part of this process. A workflow may retrieve relevant evidence but fail to combine it into a complete answer. Changing the retrieval strategy along with the integration step may disturb work that was already useful. Conversely, retaining an agent's configuration does not establish that its previous result remains applicable, since a revision may change the information it receives. Workflow evolution therefore needs to consider both the scope of a revision and its consequences for completed computations.

Inheritance gives evolution a memory, while natural selection favors beneficial variations and acts against harmful ones~\citep{darwin1859origin}. This interplay suggests that test-time MAS evolution should use execution feedback to guide change without unnecessarily losing useful work from earlier attempts. An unsuccessful attempt can still be informative, as its trace may reveal useful partial results alongside missing evidence, redundant work, or coordination failures. Such feedback offers a basis for cumulative adaptation, in which each attempt builds on what earlier executions have revealed. The aim is not to preserve every earlier choice, since some choices must be revised or abandoned, but to use this experience to guide subsequent changes. Under a finite test-time budget, both the quality of those changes and the cost of evaluating them matter. We therefore ask: \emph{How can a MAS use execution feedback to improve task performance while controlling the inference cost of test-time evolution?}

To answer this question, we propose \textbf{Inherit-MAS}, a test-time evolution loop with explicit workflow and execution inheritance. A meta-model synthesizes an initial workflow from the task and its interface, specifying agent roles, prompts, tools, and communication links. Each valid candidate is executed, and a separately prompted judge scores its output and trace and provides node-level feedback. In each ordinary refinement round, \emph{workflow inheritance} starts from the latest completed candidate (the incumbent). Guided by the critique, the meta-model may discard removable nodes judged unhelpful, then applies one validated edit intended to address a reported deficiency. The rest of the kept workflow is inherited unchanged. During execution, \emph{execution inheritance} reuses an eligible node's stored result only when its complete resolved request (including upstream results and the execution-context fingerprint) matches a verified record. All other nodes execute live. After the attempt budget is exhausted, the ranking rule returns the highest-ranked completed candidate's stored output.

We evaluate Inherit-MAS on WorkBench~\citep{styles2024workbench} and HotpotQA FullWiki~\citep{yang2018hotpotqa}. The paper makes three contributions:
\begin{itemize}
 \item We formulate test-time MAS evolution as explicit workflow inheritance. After initial synthesis, each candidate starts from the latest candidate, keeps the nodes that a node-level critique identifies as useful, and changes the kept workflow through one validated edit directed at the diagnosed deficiency.
 \item We introduce conservative change localization and request-verified execution inheritance, allowing eligible model and read-only tool results to be used directly across related candidates without altering the candidate being evaluated.
 \item On WorkBench and HotpotQA FullWiki benchmarks, Inherit-MAS outperforms EvoAgent, EvoMAS, and TacoMAS on overall completion and joint F1 scores while using less inference than EvoMAS and TacoMAS. We also show that, compared with independent reruns with execution inheritance disabled, execution inheritance avoids 29.1\% and 34.6\% of the live worker inference on the two benchmarks and 5.3\% and 18.1\% of total tokens.
\end{itemize}

\section{Background and Problem Formulation}

\begin{wrapfigure}[16]{r}{0.46\textwidth}
\centering
\includegraphics[width=\linewidth]{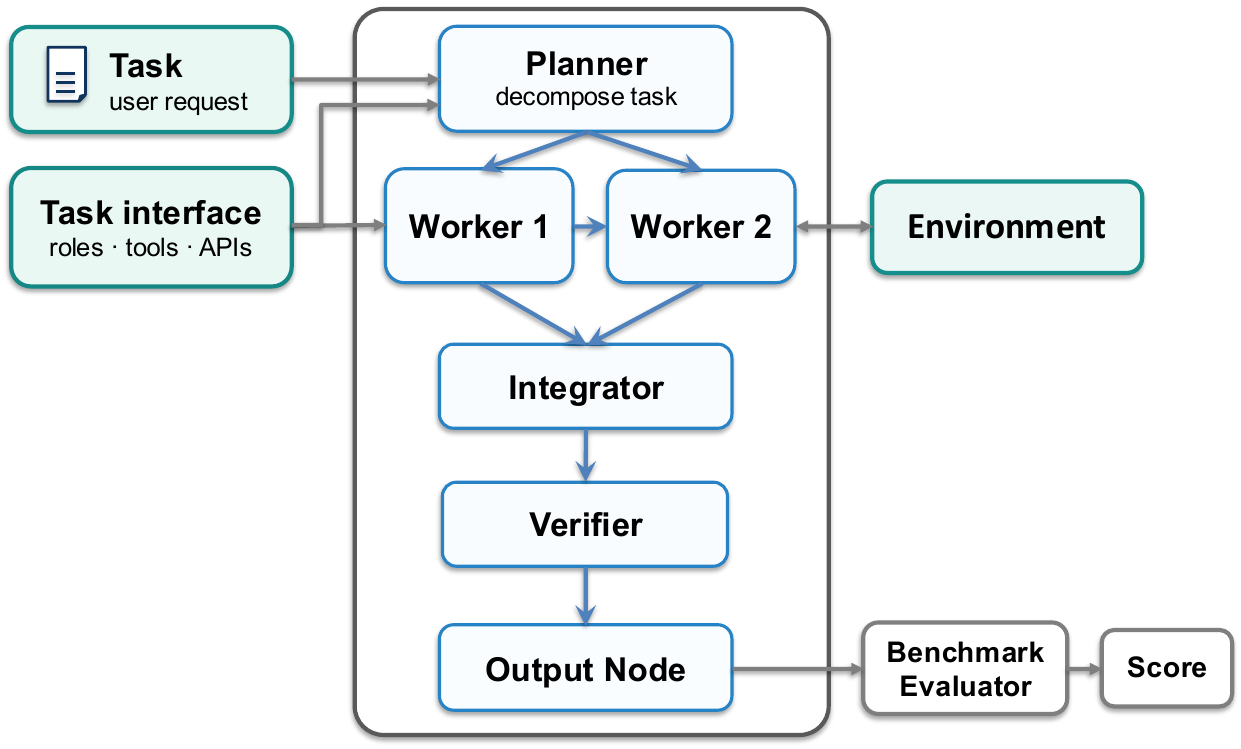}
\caption{Illustrative multi-agent workflow. The task, its interface, and the environment (green) are external to the workflow (outlined box). The benchmark evaluator scores the output.}
\label{fig:mas-overview}
\end{wrapfigure}

\begin{samepage}
\paragraph{Multi-Agent Systems.} An LLM-based multi-agent system (MAS) coordinates agents with complementary roles to divide a task and exchange intermediate results. \Cref{fig:mas-overview} illustrates one workflow in which a planner guides workers that interact with the environment, an integrator combines their findings, and a verifier checks the result before the final output is produced.
\par\end{samepage}

\Needspace{20\baselineskip}

Given a task $x$ and an interface $\mathcal I$ specifying allowed roles, tools, input/output formats, and the execution environment, a workflow-based MAS can be represented by a directed graph $G=(V,E)$. Here $V$ is the node set and $E$ is the set of directed edges carrying intermediate results between nodes. A node $v\in V$ is an agent or workflow operation with a role, prompt, tool set, decoding configuration, and declared inputs and outputs. Executing $G$ yields an output $y$, an observable trace $\tau$ of messages and tool interactions, and a cost ledger $c$. Task performance is measured by a utility function $U$, which assigns a score $U(y)$ to the answer or resulting environment state.

\paragraph{Evolving Multi-Agent Systems.} An evolving MAS uses execution feedback to adapt agent configurations and interactions rather than keeping its workflow fixed. For task-specific evolution, let $K$ be the number of candidate attempts and let $k\in\{0,\ldots,K-1\}$ index them in construction order. The workflow proposed at index $k$ is denoted by $G_k$. Executing a valid $G_k$ yields its output $y_k$, trace $\tau_k$, and cost ledger $c_k$. Later candidates may revise an earlier workflow or recombine components from a population, using feedback such as execution outcomes, environment rewards, or external evaluations. The final output may be selected from one candidate or assembled from several candidates' outputs. In the single-candidate case, $k^*$ denotes the selected candidate's index and $y_{k^*}$ its output. Evolution aims to improve task performance under the available computational and interaction budgets.

\section{Related Work}
\paragraph{Multi-Agent Systems.} Multi-agent systems coordinate role-specialized language-model agents through communication, tool use, and structured handoffs. ReAct~\citep{yao2023react} established an action-observation loop for tool-using language models, CAMEL~\citep{li2023camel} studied cooperation through role-playing agents, AutoGen~\citep{wu2024autogen} made inter-agent conversations programmable, and MetaGPT~\citep{hong2024metagpt} organized specialized roles around procedural workflows. Automatic design methods search agent programs through meta-search~\citep{hu2024adas}, optimize graph prompts and connectivity~\citep{zhuge2024gptswarm}, and construct natural-language or executable workflows~\citep{li2024autoflow,zhang2024aflow}. AgentSquare~\citep{shang2024agentsquare} searches modular agent designs. DyLAN~\citep{liu2024dylan} and G-Designer~\citep{zhang2025gdesigner} adapt team composition or communication topology, while DarkForest~\citep{li2026darkforest} passes calibrated, policy-approved evidence to a coordinator. MaAS~\citep{zhang2025maas}, FlowReasoner~\citep{gao2025flowreasoner}, and MAS-GPT~\citep{ye2025masgpt} construct query-dependent systems. These methods establish the roles, tools, prompts, and graph structures that an evolving MAS can refine.

\paragraph{Evolving Multi-Agent Systems.} Evolving MAS methods use execution feedback to revise agent configurations and communication structures. EvoAgent~\citep{yuan2024evoagent} expands an initial expert into a multi-agent system through evolutionary operations. EvoMAS~\citep{hu2026evomas} evolves configurations through mutation and crossover while retaining promising configurations and accumulated experience. TacoMAS~\citep{xu2026tacomas} couples fast updates to agent capabilities with slower changes to the agent population and communication graph. Agentic Neural Network~\citep{ma2026ann} refines layered agent teams from textual feedback, and Evoflux~\citep{bhandari2026evoflux} repairs tool-workflow graphs through schema-validated edits. Skill-centric approaches evolve reusable orchestration knowledge and execution procedures~\citep{lin2026skillmas,pan2026skillmas}. However, workflow updates can span several components and affect downstream computations. Retaining configurations or prior outputs alone does not establish which stored results remain valid after such changes. Inherit-MAS combines explicit selection of workflow components to retain, validated refinement of the retained workflow, and inheritance of stored execution results when the complete request and execution context match. Appendix~\ref{sec:positioning} provides detailed comparisons with workflow design, evolution, and execution-reuse methods.

\section{Method}
\label{sec:method}

\subsection{Overview}
\begin{figure}[t]
 \centering
 \includegraphics[width=\textwidth]{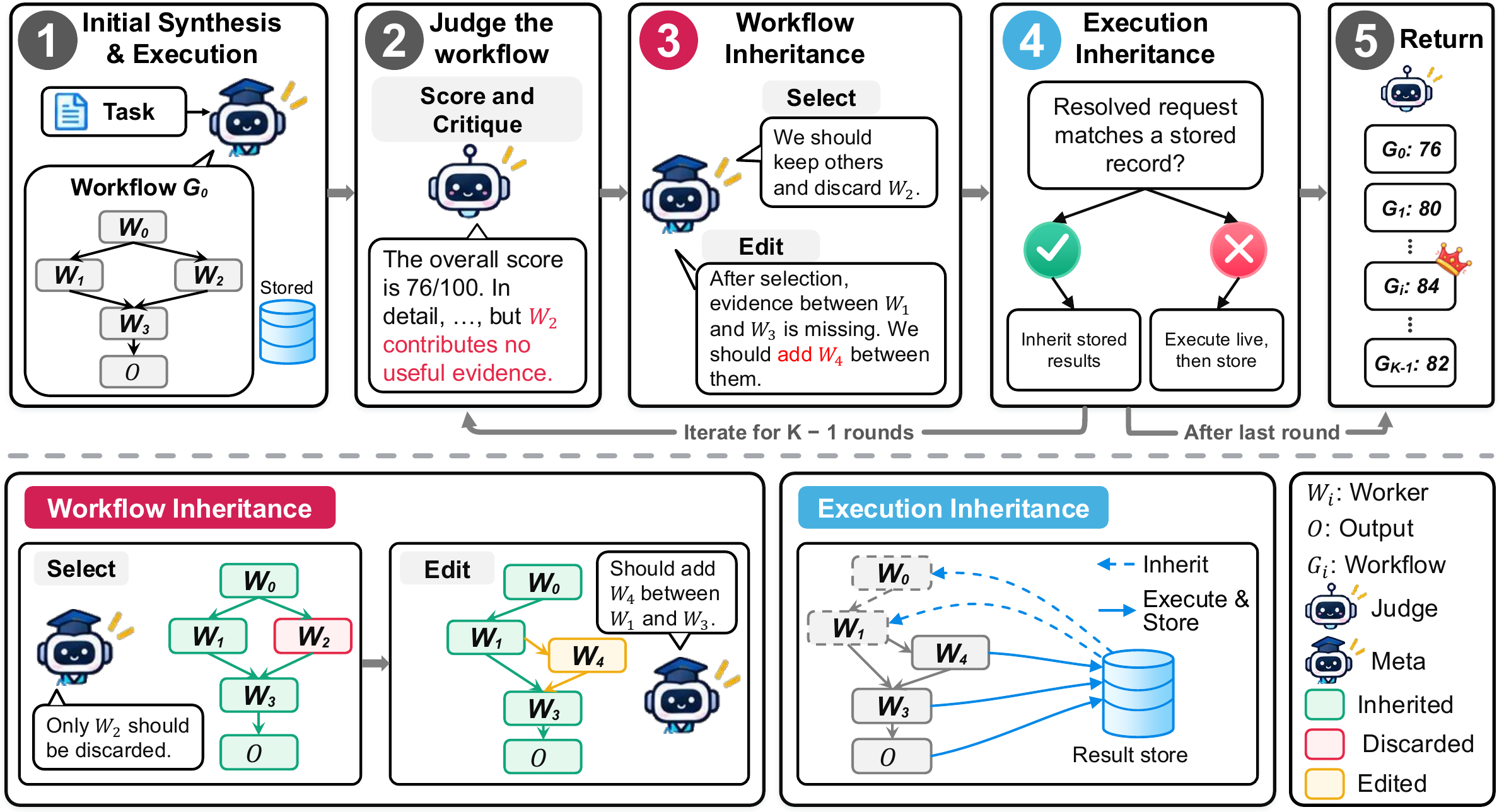}
 \caption{Overview of Inherit-MAS. Top: the ordinary path from initial synthesis and execution (1) to judgment (2), workflow inheritance (3), and execution inheritance (4), followed by final candidate ranking (5). The $K-1$ refinement rounds shown assume a successful initial attempt and omit invalid attempts and synthesis fallbacks. Bottom left: Select may discard removable nodes judged unhelpful; Edit changes the kept workflow within one declared scope. Bottom right: an eligible node inherits a result only when its complete resolved request matches a verified record. Otherwise, it executes live and stores its result if eligible.}
 \label{fig:method}
\end{figure}

Inherit-MAS searches over directed acyclic workflows $G_k=(V_k,E_k)$ within a budget of $K$ candidate attempts. \Cref{fig:method} shows the ordinary execution path. Step~1 synthesizes a valid initial workflow and executes it, storing eligible node results in an initially empty task-local result store. Step~2 uses a separately prompted judge to assess the output and trace and provide node-level feedback. In step~3, \emph{workflow inheritance} refines the latest completed candidate (the incumbent): \emph{Select} may discard removable nodes judged unhelpful, and \emph{Edit} applies one validated change to the kept workflow, intended to address a reported deficiency. Step~4 executes the new candidate with \emph{execution inheritance}: eligible nodes inherit stored results only when their complete resolved requests match verified records; all other nodes execute live. Execution is followed by judgment before the next attempt. After the attempt budget is exhausted, step~5 returns the output of the highest-ranked completed candidate. Invalid attempts and synthesis fallbacks are handled as specified below. 

\subsection{Initial Synthesis and Judgment}
The first attempt has no incumbent: the meta-model's synthesis procedure $\mathcal S$ proposes $G_0=\mathcal S(x,\mathcal I)$ from the task and its interface. A valid workflow is executed to produce $(y_k,\tau_k,c_k)$. The separately prompted judge $\mathcal J$ then returns $(q_k,\gamma_k)=\mathcal J(x,G_k,y_k,\tau_k)$, where $q_k=(q_k^{\mathrm{cmp}},q_k^{\mathrm{cov}},q_k^{\mathrm{grd}},q_k^{\mathrm{all}})\in[0,100]^4$ contains completion likelihood, requirement coverage, artifact grounding, and overall quality, and $\gamma_k$ contains the textual critique and keep/fix recommendations. A candidate is \emph{completed} once its execution and a valid judge response have been recorded. Because the judge receives the workflow and recorded node outputs, it is instructed to identify useful and unhelpful contributions when the trace supports such attribution (step~2 in \Cref{fig:method}).

\subsection{Workflow Inheritance}
\label{sec:evolution}
Workflow inheritance governs ordinary refinement from the most recently executed and judged candidate. It has two stages: \emph{Select} chooses which nodes to retain, and \emph{Edit} modifies the kept workflow. Select may discard several nodes, so the one-edit restriction applies to the kept workflow, not to the entire difference between parent and child. Candidate-level ranking is a separate operation described in \Cref{sec:final-selection}.

\paragraph{Select: inherit or discard nodes.}
In an ordinary refinement attempt $k>0$, let $p(k)<k$ denote the index of the latest completed candidate, whose workflow $G_{p(k)}$ is the incumbent. Let $\mathcal R(G)$ contain the nodes whose individual removal leaves $G$ valid under $\mathcal I$. The output node and nodes needed to satisfy required input constraints are excluded. The meta-model's node selector $\mathcal P_{\mathrm{sel}}$ is instructed to discard the node only when the feedback or recorded output indicates that it contributes nothing useful or harms the result. It proposes a discard set $\Delta^-_k$, and pruning removes those nodes and their incident edges:
\begin{equation}
\begin{aligned}
\Delta^-_k&=\mathcal P_{\mathrm{sel}}\!\left(x,G_{p(k)},\tau_{p(k)},q_{p(k)},\gamma_{p(k)}\right),\\
G'_k&=\mathsf{Prune}(G_{p(k)},\Delta^-_k).
\end{aligned}
\end{equation}
The selection is accepted only if $\Delta^-_k\subseteq\mathcal R(G_{p(k)})$ and $\mathsf{Valid}(G'_k,\mathcal I)=1$, where $\mathsf{Valid}$ is the deterministic workflow validator. Checking the combined discard set is necessary because individually removable nodes need not be jointly removable. If $\mathcal R(G_{p(k)})=\varnothing$, no node can be removed individually while preserving validity, so selection is skipped and $G'_k=G_{p(k)}$. Every kept node retains its role, prompt, assigned subtask, tools, and remaining inputs.

\paragraph{Edit: propose and validate.}
Let $\mathcal A(G)$ denote the admissible edits for graph $G$, $\mathcal P$ the proposer, and $\mathcal T$ the deterministic edit applicator. The implementation exposes a finite menu of operations and targets. The proposer chooses one entry and supplies any required replacement text or node instructions. It receives the incumbent's execution trace and judge feedback but edits the kept workflow:
\begin{equation}
\begin{aligned}
a_k&=\mathcal P\!\left(x,G'_k,\tau_{p(k)},q_{p(k)},\gamma_{p(k)},\mathcal A(G'_k)\right),\\
G_k&=\mathcal T(G'_k,a_k),\quad a_k\in\mathcal A(G'_k),\\
&\hspace{1em}\mathsf{Valid}(G_k,\mathcal I)=1.
\end{aligned}
\end{equation}
The transition is accepted only after validation. Cycles, incompatible input/output connections, unauthorized tools, and other violations of $\mathcal I$ are rejected. $\mathcal T$ changes only fields within the selected edit's declared scope, preserving the rest of the kept workflow. This guarantees structural validity and scoped modification, not that the edit repairs the diagnosed problem or improves task performance. Unsuccessful attempts consume a candidate slot and remain in failure and cost accounting. The bounded response-repair procedure is specified in Appendix~\ref{sec:controller}.

An edit may change a node's prompt, assigned subtask, or tool access; add or remove a node or edge; reorder inputs; or insert a node between connected nodes, as with $W_4$ between $W_1$ and $W_3$ in \Cref{fig:method}. The insertion adds a path through the new node while retaining the original edge. Appendix~\ref{sec:workflow-rules} gives the edit family and validator checks. When no candidate has yet completed, the next attempt repeats initial synthesis instead of refinement. With completed candidates available, the final attempt uses workflow resynthesis if none has a valid output, or if the highest overall-quality score $q^{\mathrm{all}}$ is at most one point above that of the first completed candidate. This restart prompt receives the highest-ranked prior workflow and its judge feedback and requests a meaningfully different valid workflow. It does not require component preservation and is treated as a transition without workflow inheritance. Algorithm~\ref{alg:method} gives the complete controller.

\subsection{Execution Inheritance}
\label{sec:inheritance}
Change localization identifies where requests may differ, while execution inheritance decides whether a previous result can be used. A hit requires an eligible node, an exact match of the complete resolved request, and a record that passes integrity verification. A node inside the affected region can still inherit a result if its current request matches one stored earlier.

\paragraph{Localize: identify possible downstream effects.}
After Select and Edit have been validated, localization compares $G_k$ with the original incumbent $G_{p(k)}$, accounting for both pruning and editing. Let $\Delta_k=\mathsf{Touch}(G_{p(k)},G_k)$ contain every new node and every retained node whose local specification, ordered incoming wiring, or effective execution settings change. This includes any retained node that loses an input through pruning and any node whose tool or retrieval allowance changes (Appendix~\ref{sec:runtime-limits}). The conservative affected region is
\begin{equation}
D_k=\Delta_k\cup\operatorname{Desc}_{G_k}(\Delta_k),
\label{eq:affected-region}
\end{equation}
where $\operatorname{Desc}_{G}(S)$ denotes nodes reachable from $S$ by one or more directed edges. For synthesis without a parent, $D_k=V_k$. This partition separates structurally unchanged nodes $V_k\setminus D_k$ from potentially affected nodes $D_k$. It does not imply that the behavioral effect is small, nor does it authorize inheritance. Both regions require the complete request verification described next.

\paragraph{Verify and inherit.}
For node $v\in V_k$, the \emph{resolved request} contains everything needed to execute it, including its current specification, ordered upstream records, and execution context. We form the request and its lookup key as follows.
\begin{equation}
\begin{aligned}
\phi_k&=\mathsf{Fingerprint}(e_k),\\
r_{k,v}&=\mathsf{Resolve}\!\left(
x,v,[(\pi_i,w_{k,u_i})]_{i=1}^{d_v},\phi_k\right),\\
\kappa_{k,v}&=h(r_{k,v}).
\end{aligned}
\label{eq:request}
\end{equation}
Here $d_v$ is the number of upstream inputs. Each input has a label $\pi_i$ and a record $w_{k,u_i}$ from source node $u_i$. These records include outputs, metadata, and exposed tool observations or errors. $\mathsf{Resolve}$ includes the actual messages, model and decoding settings, allowed tools, and execution limits. The context $e_k$ covers the runtime and parser, initial environment state, and fixed tool or retrieval service. Its fingerprint $\phi_k$ reflects their contents and versions, regardless of which environment copy is used. The hash $h$ applies SHA-256 to the canonically serialized request. The node's own tool observations arise after lookup and are stored with its result, not included in the key.

The task-local store $\mathcal M$ maps request keys to records containing output $o$, tool observations and artifacts $A$, original token usage $t$, and execution errors $\epsilon$. All candidates share this store. A node inherits a result only if it is eligible and a matching record passes verification,
\begin{equation}
z_{k,v}=\mathbf{1}\{\mathsf{Eligible}(v,e_k)\ \land\
\mathsf{ValidRecord}(\mathcal M,\kappa_{k,v})=1\},
\label{eq:hit}
\end{equation}
where $\mathsf{Eligible}$ checks whether inheritance is permitted. $\mathsf{ValidRecord}$ checks the record's format, request key, and content hash. It returns one for a verified match and zero when no record exists. A malformed or inconsistent record stops execution instead of being treated as a miss.

Language-model nodes without state-changing tools and deterministic read-only operations can be eligible. State-changing operations are never eligible. For stateful tasks, each candidate starts from a fresh copy of the same initial environment. Intermediate nodes can only read it, and a designated sink performs every state-changing action live. After verification, the node either inherits the stored result or executes live,
\begin{equation}
(o,A,t,\epsilon)_{k,v}=\begin{cases}
\mathcal M[\kappa_{k,v}], & z_{k,v}=1,\\
\mathsf{Exec}(v,r_{k,v},e_k), & z_{k,v}=0.
\end{cases}
\label{eq:inheritance}
\end{equation}
Eligible live results are stored without later modification. The downstream record $w_{k,v}$ contains the node's metadata, output, exposed artifacts, and errors, but not token usage. Inherited results restore the same evidence for downstream nodes and the judge. Inherited errors remain errors.

A hit uses no live worker tokens. Its original usage is recorded separately as an inherited token equivalent. Synthesis, selection, refinement, and judging remain live. Inheritance restores a recorded result without assuming that a fresh remote-model call would reproduce it. Appendices~\ref{sec:ledger} and~\ref{sec:store-footprint} give the accounting details and deterministic-runtime checks.

\subsection{Ranking and Return}
\label{sec:final-selection}
Let $\mathcal C\subseteq\{0,\ldots,K-1\}$ be the indices of completed candidates. For $k\in\mathcal C$, let $b_k$ count recorded node execution errors plus one if the final output is invalid under the benchmark adapter. The online ranking key and returned candidate are
\begin{equation}
\begin{aligned}
s_k&=\left(-b_k,\mathbf{1}\{y_k\text{ is valid}\},q_k^{\mathrm{all}},q_k^{\mathrm{cov}},q_k^{\mathrm{grd}},-k\right),\\
k^*&=\arg\max_{i\in\mathcal C}^{\mathrm{prio}}s_i.
\end{aligned}
\end{equation}
Here $\mathbf{1}$ is the binary indicator, and priority comparison proceeds from left to right; the first differing component determines the preferred candidate. A tie on all preceding components favors the earlier candidate. Cost is excluded from the key. If $\mathcal C$ is empty, no candidate output is returned and the task is recorded as a failure. Otherwise, step~5 returns $y_{k^*}$ without re-executing the workflow. Retaining all completed candidates prevents a lower-ranked child from displacing its parent in the final selection, but this ordering does not guarantee improvement in the unavailable benchmark utility $U$.

\section{Experiments}
\label{sec:results}

\subsection{Experimental Setup}
\label{subsec:setup}
\paragraph{Baselines.} We compare our proposed \textbf{Inherit-MAS} with four baselines. \textbf{Single ReAct}~\citep{yao2023react} is one agent with the complete public tool interface and no meta-model, evolution, or execution inheritance. We report it as a non-MAS reference. \textbf{EvoAgent}~\citep{yuan2024evoagent} grows a population from an initial expert agent. \textbf{EvoMAS}~\citep{hu2026evomas} evolves structured MAS configurations with execution traces and persistent cross-task experience. \textbf{TacoMAS}~\citep{xu2026tacomas} couples fast capability updates with slower topology evolution.

\paragraph{Benchmarks.} We use two benchmarks for the evaluation: \textbf{WorkBench}~\citep{styles2024workbench} and \textbf{HotpotQA}~\citep{yang2018hotpotqa}. WorkBench contains workplace tasks over email, calendar, project management, CRM, and analytics databases, with outcome-centric evaluation of the final state. HotpotQA requires multi-document reasoning and sentence-level supporting facts. Inherit-MAS, Single ReAct, TacoMAS, and the Qwen EvoMAS row share a budget of four calls per candidate to the same fixed BM25 index, returning five documents per call and allocated across a candidate's retrieval nodes; EvoAgent and the GPT-4o-mini EvoMAS row receive four calls per agent (Appendix~\ref{sec:baseline-fidelity}).

\paragraph{Implementation.} Inherit-MAS uses GPT-4o-mini workers~\citep{openai2024gpt4omini} and GPT-5.4-mini~\citep{openai2026gpt54mini} for synthesis/refinement and the separately prompted judge, all at temperature zero. Each trajectory contains an initial workflow and four subsequent candidates, with one Select decision and one validated edit ordinarily applied per refinement round. Single ReAct uses the same worker model. For WorkBench, every candidate runs from a fresh environment copy and a deterministic executor applies its proposed write batch.

All MAS baselines use released implementations with benchmark-specific adaptations. EvoAgent uses GPT-4o-mini; EvoMAS and TacoMAS use GPT-4o-mini workers with GPT-5.4-mini meta-models and judges. The second backbone configuration uses Qwen3-32B~\citep{yang2025qwen3} workers in non-thinking mode for all five systems. Meta-models and judges remain GPT-5.4-mini, while EvoAgent uses Qwen3-32B throughout. Appendices~\ref{sec:baseline-fidelity} and~\ref{sec:qwen-settings} provide the code versions, benchmark adaptations, model-routing details, and decoding settings.

\paragraph{Evaluation.} All systems are evaluated on the same fixed set of tasks, and the code and configurations of every system were fixed before the final runs. During a run, the meta-model and judge never observe reference actions, answers, supporting facts, or official metrics, which are used only for post-run scoring (Appendix~\ref{sec:search-inputs}). Failures and invalid outputs remain in the denominator. On WorkBench, the benchmark evaluator scores the final environment state, and the primary metric is official completion. On HotpotQA, the primary metric is official joint F1, which requires the returned answer together with sentence-level supporting facts in the required structured form, and we also report answer and supporting-fact F1. For cost, every run records the token usage of every model call, which we split into worker inference and meta-model and judge inference for synthesis, node selection, refinement, and judging, and inherited results are reported separately as token equivalents. External systems run under their documented configurations. 

\begin{table}[t]
\centering
\caption{WorkBench completion percentages by worker backbone and domain. Bold marks the best MAS result within each backbone, excluding the Single ReAct reference. The shaded Overall column is the primary metric, computed over all evaluation tasks (Appendix~\ref{sec:reproducibility}).}
\label{tab:workbench}
\scriptsize
\setlength{\tabcolsep}{4.2pt}
\resizebox{.9\textwidth}{!}{%
\begin{tabular}{@{}lcccccc>{\columncolor{lightgray}}c}
\toprule
System & Analytics & Calendar & CRM & Email & Project mgmt. & Multi-domain & \textbf{Overall}\\
\midrule
\multicolumn{8}{c}{Workers: GPT-4o-mini}\\
\midrule
Single ReAct (reference) & 35.0 & 50.0 & 20.0 & 35.0 & 20.0 & 60.0 & 41.5\\
\paddedhdashline
EvoAgent & 55.0 & 40.0 & \textbf{20.0} & 25.0 & 20.0 & 35.0 & 33.8\\
EvoMAS & 25.0 & 35.0 & \textbf{20.0} & 35.0 & 20.0 & 37.5 & 30.8\\
TacoMAS & 20.0 & 55.0 & \textbf{20.0} & 5.0 & 20.0 & 57.5 & 34.6\\
\paddedhdashline
\textbf{Inherit-MAS} & \textbf{60.0} & \textbf{75.0} & \textbf{20.0} & \textbf{45.0} & \textbf{35.0} & \textbf{67.5} & \textbf{55.4}\\
\midrule
\multicolumn{8}{c}{Workers: Qwen3-32B}\\
\midrule
Single ReAct (reference) & 65.0 & 55.0 & 20.0 & 35.0 & 25.0 & 55.0 & 46.2\\
\paddedhdashline
EvoAgent & 25.0 & 30.0 & 20.0 & 20.0 & 20.0 & 40.0 & 28.5\\
EvoMAS & 20.0 & 35.0 & 20.0 & 30.0 & 15.0 & \textbf{52.5} & 33.1\\
TacoMAS & 35.0 & \textbf{60.0} & \textbf{40.0} & 0.0 & \textbf{25.0} & 42.5 & 34.6\\
\paddedhdashline
\textbf{Inherit-MAS} & \textbf{75.0} & 50.0 & 20.0 & \textbf{40.0} & 20.0 & \textbf{52.5} & \textbf{46.2}\\
\bottomrule
\end{tabular}
}
\end{table}

\subsection{Main Results}
Inherit-MAS outperforms the evolving-MAS baselines on both benchmarks with both worker backbones (\Cref{tab:workbench,tab:hotpot}). On WorkBench with GPT-4o-mini workers, it is 13.8 points above Single ReAct and 20.8 points above TacoMAS (the best evolving-MAS baseline), while every evolving-MAS baseline trails Single ReAct. With Qwen3-32B workers, it matches Single ReAct at 46.2\% and exceeds the best evolving-MAS baseline by 11.5 points. 
On HotpotQA, joint F1 couples the answer with sentence-level citations. With GPT-4o-mini workers, Inherit-MAS is 3.7 points above Single ReAct and 1.9 points above TacoMAS, the closest competitor. It has the highest answer F1 of all systems, while its supporting-fact F1 stays at Single ReAct's level, and TacoMAS has the best supporting facts but weaker answers. The same pattern holds with Qwen3-32B workers, where Inherit-MAS is 5.6 points above Single ReAct and 2.8 points above TacoMAS, again with the highest answer F1, while TacoMAS retains the best supporting facts.

\begin{table}[t]
\centering
\caption{HotpotQA FullWiki F1 scores (\%) by worker backbone. Bold marks the best MAS result within each backbone. SP F1 is supporting-fact F1. The shaded Joint F1 column is the primary metric, the official combination of answer and supporting-fact scores (Appendix~\ref{sec:reproducibility}).}
\label{tab:hotpot}
\footnotesize
\setlength{\tabcolsep}{8pt}
\begin{tabular}{@{}lcc>{\columncolor{lightgray}}c}
\toprule
System & Answer F1 & SP F1 & \textbf{Joint F1}\\
\midrule
\multicolumn{4}{c}{Workers: GPT-4o-mini}\\
\midrule
Single ReAct (reference) & 69.5 & 61.9 & 46.0\\
\paddedhdashline
EvoAgent & 69.8 & 47.4 & 34.8\\
EvoMAS & 64.5 & 54.7 & 39.2\\
TacoMAS & 66.8 & \textbf{64.8} & 47.8\\
\paddedhdashline
\textbf{Inherit-MAS} & \textbf{73.3} & 61.7 & \textbf{49.7}\\
\midrule
\multicolumn{4}{c}{Workers: Qwen3-32B}\\
\midrule
Single ReAct (reference) & 63.0 & 51.7 & 37.1\\
\paddedhdashline
EvoAgent & 61.5 & 45.7 & 32.3\\
EvoMAS & 61.1 & 56.4 & 39.4\\
TacoMAS & 56.8 & \textbf{60.0} & 39.9\\
\paddedhdashline
\textbf{Inherit-MAS} & \textbf{65.3} & 58.5 & \textbf{42.7}\\
\bottomrule
\end{tabular}
\end{table}

Across the four benchmark--backbone settings, Inherit-MAS is the only evolving-MAS system that matches or exceeds Single ReAct in every setting, and its margin over the other evolving-MAS systems is widest on WorkBench, where completion is scored on the final environment state. In each ordinary refinement round, Inherit-MAS inherits the nodes of the latest candidate that the judge finds useful and changes the kept workflow by the validated edit. Inherit-MAS reaches these scores with fewer tokens than EvoMAS and TacoMAS, which use 16.8$\times$ and 22.9$\times$ its total on WorkBench and 13.7$\times$ and 2.1$\times$ on HotpotQA under each system's own configuration (Appendix~\ref{sec:cost-appendix}).

\subsection{Performance across Refinement Rounds}
\begin{figure}[t]
\centering
\begin{minipage}[t]{0.48\textwidth}
\centering
\includegraphics[width=\linewidth]{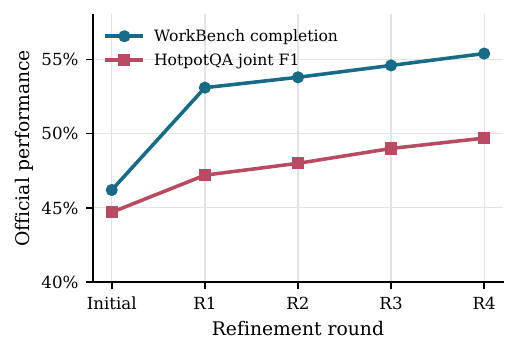}
\caption{Official performance of the returned workflow after initial synthesis (Initial) and after each refinement round (R1 to R4, counted by candidate index). The R4 points are the reported final results.}
\label{fig:refinement-rounds}
\end{minipage}\hfill
\begin{minipage}[t]{0.48\textwidth}
\centering
\includegraphics[width=\linewidth]{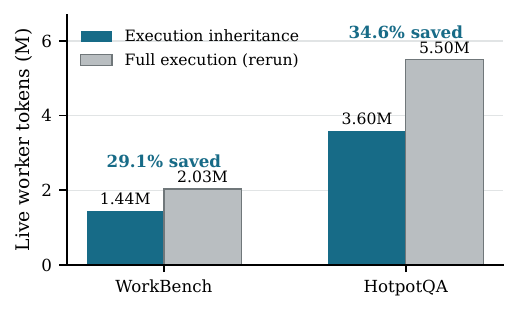}
\caption{Live worker tokens with execution inheritance (original runs) and with execution inheritance disabled (independent reruns of the same controller on the same tasks, averaged over two reruns on WorkBench).}
\label{fig:full-eval-cost}
\end{minipage}
\end{figure}

After each task's evolution finishes, we score offline the workflow that Inherit-MAS would have returned after initial synthesis (Initial) and after each refinement round (R1 to R4), using the official metric (\Cref{fig:refinement-rounds}). Both curves rise at every round, by 5.0 points in total on HotpotQA and 9.2 points on WorkBench, where the first round contributes 6.9 points. These gains show that the evolution process yields progressively better performance.

\subsection{Execution Inheritance and Token Cost}
We rerun Inherit-MAS on the same tasks with execution inheritance disabled, keeping the controller, judge, edit menu, and retrieval budget unchanged. Every node executes live, and the reruns are independent trajectories.

From \Cref{fig:full-eval-cost}, full execution spends 2.03M worker tokens on WorkBench and 5.50M on HotpotQA, against 1.44M and 3.60M with execution inheritance, so execution inheritance saves 29.1\% and 34.6\% of worker tokens. Meta-model and judge inference stays live in both settings, so total token savings are 5.3\% and 18.1\%, smaller on WorkBench where worker calls are a smaller share of the budget. The savings come from requests that an edit leaves unchanged, so within the original trajectories they are concentrated in prompt edits, which inherit 47.3\% and 75.0\% of worker-token equivalents (\Cref{tab:edit-transitions}). Full execution achieves 55.4\% completion in the WorkBench rerun, the same as execution inheritance.


\section{Conclusion}

Inherit-MAS makes inheritance explicit in test-time multi-agent system evolution by keeping an incumbent's useful nodes, applying one validated, diagnosis-targeted edit to the kept workflow, and inheriting stored results for unchanged requests. On WorkBench and HotpotQA benchmarks, it achieves higher primary-metric scores than the evolving-MAS baselines with both GPT-4o-mini and Qwen3-32B workers. In the GPT-4o-mini experiments, execution inheritance reduces worker-token usage by 29.1\% and 34.6\% relative to independent full-execution reruns, corresponding to total token savings of 5.3\% and 18.1\%.

\clearpage
\bibliographystyle{plainnat}
\bibliography{references}

@book{darwin1859origin,
  title={On the Origin of Species by Means of Natural Selection, or the Preservation of Favoured Races in the Struggle for Life},
  author={Darwin, Charles},
  publisher={John Murray},
  address={London},
  edition={First},
  year={1859},
  url={https://darwin-online.org.uk/converted/published/1859_Origin_F373/1859_Origin_F373.html}
}

@inproceedings{yao2023react,
  title={{ReAct}: Synergizing Reasoning and Acting in Language Models},
  author={Yao, Shunyu and Zhao, Jeffrey and Yu, Dian and Du, Nan and Shafran, Izhak and Narasimhan, Karthik and Cao, Yuan},
  booktitle={International Conference on Learning Representations},
  year={2023},
  url={https://arxiv.org/abs/2210.03629}
}

@inproceedings{styles2024workbench,
  title={{WorkBench}: a Benchmark Dataset for Agents in a Realistic Workplace Setting},
  author={Styles, Olly and Miller, Sam and Cerda-Mardini, Patricio and Guha, Tanaya and Sanchez, Victor and Vidgen, Bertie},
  booktitle={First Conference on Language Modeling},
  year={2024},
  url={https://openreview.net/forum?id=4HNAwZFDcH}
}

@inproceedings{yang2018hotpotqa,
  title={{HotpotQA}: A Dataset for Diverse, Explainable Multi-hop Question Answering},
  author={Yang, Zhilin and Qi, Peng and Zhang, Saizheng and Bengio, Yoshua and Cohen, William W. and Salakhutdinov, Ruslan and Manning, Christopher D.},
  booktitle={Proceedings of the 2018 Conference on Empirical Methods in Natural Language Processing},
  pages={2369--2380},
  year={2018},
  doi={10.18653/v1/D18-1259}
}

@article{yang2025qwen3,
  title={{Qwen3} Technical Report},
  author={Yang, An and Li, Anfeng and Yang, Baosong and Zhang, Beichen and Hui, Binyuan and Zheng, Bo and Yu, Bowen and others},
  journal={arXiv preprint arXiv:2505.09388},
  year={2025},
  url={https://arxiv.org/abs/2505.09388}
}

@inproceedings{hu2026evomas,
  title={{EvoMAS}: Evolutionary Generation of Multi-Agent Systems},
  author={Hu, Yuntong and Zhang, Yuting and Trager, Matthew and Zhang, Yi and Yang, Shuo and Xia, Wei and Soatto, Stefano},
  booktitle={International Conference on Machine Learning},
  year={2026},
  url={https://arxiv.org/abs/2602.06511}
}

@article{xu2026tacomas,
  title={{TacoMAS}: Test-Time Co-Evolution of Topology and Capability in {LLM}-based Multi-Agent Systems},
  author={Xu, Chen and Hu, Yicheng and Wang, Ruizi and Lin, Xinyu and Wang, Wenjie and Liu, Dongrui and Feng, Fuli},
  journal={arXiv preprint arXiv:2605.09539},
  year={2026},
  url={https://arxiv.org/abs/2605.09539}
}

@inproceedings{yuan2024evoagent,
  title={{EvoAgent}: Towards Automatic Multi-Agent Generation via Evolutionary Algorithms},
  author={Yuan, Siyu and Song, Kaitao and Chen, Jiangjie and Tan, Xu and Li, Dongsheng and Yang, Deqing},
  booktitle={Proceedings of the 2025 Conference of the Nations of the Americas Chapter of the Association for Computational Linguistics: Human Language Technologies (Volume 1: Long Papers)},
  pages={6192--6217},
  year={2025},
  doi={10.18653/v1/2025.naacl-long.315},
  url={https://aclanthology.org/2025.naacl-long.315/}
}

@inproceedings{hu2024adas,
  title={Automated Design of Agentic Systems},
  author={Hu, Shengran and Lu, Cong and Clune, Jeff},
  booktitle={International Conference on Learning Representations},
  year={2025},
  url={https://proceedings.iclr.cc/paper_files/paper/2025/hash/36b7acf6f6010652b3f2a433774a66fe-Abstract-Conference.html}
}

@article{li2024autoflow,
  title={{AutoFlow}: Automated Workflow Generation for Large Language Model Agents},
  author={Li, Zelong and Xu, Shuyuan and Mei, Kai and Hua, Wenyue and Rama, Balaji and Raheja, Om and Wang, Hao and Zhu, He and Zhang, Yongfeng},
  journal={arXiv preprint arXiv:2407.12821},
  year={2024},
  url={https://arxiv.org/abs/2407.12821}
}

@article{lin2026skillmas,
  title={{Skill-MAS}: Evolving Meta-Skill for Automatic Multi-Agent Systems},
  author={Lin, Hehai and Yang, Qi and Qin, Chengwei},
  journal={arXiv preprint arXiv:2606.18837},
  year={2026},
  url={https://arxiv.org/abs/2606.18837}
}

@inproceedings{abhyankar2024infercept,
  title={{InferCept}: Efficient Intercept Support for Augmented Large Language Model Inference},
  author={Abhyankar, Reyna and He, Zijian and Srivatsa, Vikranth and Zhang, Hao and Zhang, Yiying},
  booktitle={Proceedings of the 41st International Conference on Machine Learning},
  series={Proceedings of Machine Learning Research},
  volume={235},
  pages={81--95},
  year={2024},
  url={https://proceedings.mlr.press/v235/abhyankar24a.html}
}

@article{pan2026skillmas,
  title={{SkillMAS}: Skill Co-Evolution with {LLM}-based Multi-Agent System},
  author={Pan, Shuai and Liu, Yixiang and Gao, Jiaye and Gao, Te and Liu, Weiwen and Lin, Jianghao and Fu, Zhihui and Wang, Jun and Zhang, Weinan and Yu, Yong},
  journal={arXiv preprint arXiv:2605.09341},
  year={2026},
  url={https://arxiv.org/abs/2605.09341}
}

@article{bhandari2026evoflux,
  title={Evoflux: Inference-Time Evolution of Executable Tool Workflows for Compact Agents},
  author={Bhandari, Kushal Raj and Yue, Ling and Ko, Ching-Yun and Patel, Dhaval and Pan, Shaowu and Chen, Pin-Yu and Gao, Jianxi},
  journal={arXiv preprint arXiv:2606.12674},
  year={2026},
  url={https://arxiv.org/abs/2606.12674}
}

@inproceedings{zhang2024aflow,
  title={{AFlow}: Automating Agentic Workflow Generation},
  author={Zhang, Jiayi and Xiang, Jinyu and Yu, Zhaoyang and Teng, Fengwei and Chen, Xionghui and Chen, Jiaqi and Zhuge, Mingchen and Cheng, Xin and Hong, Sirui and Wang, Jinlin and Zheng, Bingnan and Liu, Bang and Luo, Yuyu and Wu, Chenglin},
  booktitle={International Conference on Learning Representations},
  year={2025},
  url={https://openreview.net/forum?id=z5uVAKwmjf}
}

@inproceedings{zhang2025maas,
  title={Multi-agent Architecture Search via Agentic Supernet},
  author={Zhang, Guibin and Niu, Luyang and Fang, Junfeng and Wang, Kun and Bai, Lei and Wang, Xiang},
  booktitle={Proceedings of the 42nd International Conference on Machine Learning},
  series={Proceedings of Machine Learning Research},
  volume={267},
  pages={75834--75852},
  year={2025},
  url={https://proceedings.mlr.press/v267/zhang25bi.html}
}

@inproceedings{shang2024agentsquare,
  title={{AgentSquare}: Automatic {LLM} Agent Search in Modular Design Space},
  author={Shang, Yu and Li, Yu and Zhao, Keyu and Ma, Likai and Liu, Jiahe and Xu, Fengli and Li, Yong},
  booktitle={International Conference on Learning Representations},
  year={2025},
  url={https://arxiv.org/abs/2410.06153}
}

@article{nouri2026stepcache,
  title={{StepCache}: Step-Level Reuse with Lightweight Verification and Selective Patching for {LLM} Serving},
  author={Nouri, Azam},
  journal={arXiv preprint arXiv:2603.28795},
  year={2026},
  url={https://arxiv.org/abs/2603.28795}
}

@article{wadlom2026helium,
  title={Efficient {LLM} Serving for Agentic Workflows: A Data Systems Perspective},
  author={Wadlom, Noppanat and Shen, Junyi and Lu, Yao},
  journal={Proceedings of the ACM on Management of Data},
  volume={4},
  number={3},
  pages={1--29},
  year={2026},
  doi={10.1145/3802046}
}

@inproceedings{zhang2025plancache,
  title={Agentic Plan Caching: Test-Time Memory for Fast and Cost-Efficient {LLM} Agents},
  author={Zhang, Qizheng and Wornow, Michael and Olukotun, Kunle},
  booktitle={Advances in Neural Information Processing Systems},
  volume={38},
  year={2025},
  url={https://proceedings.neurips.cc/paper_files/paper/2025/hash/9549f7d06700f0966d5f938f1d11022a-Abstract-Conference.html}
}

@article{yuan2026flowbank,
  title={{FlowBank}: Query-Adaptive Agentic Workflows Optimization through Precompute-and-Reuse},
  author={Yuan, Lingzhi and Deng, Chenghao and Yu, Fangxu and Chakraborty, Souradip and Rostami, Mohammad and Huang, Furong},
  journal={arXiv preprint arXiv:2606.11290},
  year={2026},
  url={https://arxiv.org/abs/2606.11290}
}

@article{fareed2026speculative,
  title={Cost-Aware Speculative Execution for {LLM}-Agent Workflows: An Integrated Five-Dimension Method},
  author={Fareed, Faisal},
  journal={arXiv preprint arXiv:2606.07846},
  year={2026},
  url={https://arxiv.org/abs/2606.07846}
}

@inproceedings{wang2023selfconsistency,
  title={Self-Consistency Improves Chain of Thought Reasoning in Language Models},
  author={Wang, Xuezhi and Wei, Jason and Schuurmans, Dale and Le, Quoc V. and Chi, Ed H. and Narang, Sharan and Chowdhery, Aakanksha and Zhou, Denny},
  booktitle={International Conference on Learning Representations},
  year={2023},
  url={https://arxiv.org/abs/2203.11171}
}

@inproceedings{snell2024testtime,
  title={Scaling {LLM} Test-Time Compute Optimally Can be More Effective than Scaling Parameters for Reasoning},
  author={Snell, Charlie and Lee, Jaehoon and Xu, Kelvin and Kumar, Aviral},
  booktitle={International Conference on Learning Representations},
  year={2025},
  url={https://proceedings.iclr.cc/paper_files/paper/2025/hash/1b623663fd9b874366f3ce019fdfdd44-Abstract-Conference.html}
}

@inproceedings{shinn2023reflexion,
  title={Reflexion: Language Agents with Verbal Reinforcement Learning},
  author={Shinn, Noah and Cassano, Federico and Gopinath, Ashwin and Narasimhan, Karthik and Yao, Shunyu},
  booktitle={Advances in Neural Information Processing Systems},
  volume={36},
  year={2023},
  url={https://proceedings.neurips.cc/paper_files/paper/2023/hash/1b44b878bb782e6954cd888628510e90-Abstract-Conference.html}
}

@inproceedings{madaan2023selfrefine,
  title={Self-Refine: Iterative Refinement with Self-Feedback},
  author={Madaan, Aman and Tandon, Niket and Gupta, Prakhar and Hallinan, Skyler and Gao, Luyu and Wiegreffe, Sarah and Alon, Uri and Dziri, Nouha and Prabhumoye, Shrimai and Yang, Yiming and Gupta, Shashank and Majumder, Bodhisattwa Prasad and Hermann, Katherine and Welleck, Sean and Yazdanbakhsh, Amir and Clark, Peter},
  booktitle={Advances in Neural Information Processing Systems},
  year={2023},
  url={https://arxiv.org/abs/2303.17651}
}

@inproceedings{hammer2015incremental,
  title={Incremental Computation with Names},
  author={Hammer, Matthew A. and Dunfield, Jana and Headley, Kyle and Labich, Nicholas and Foster, Jeffrey S. and Hicks, Michael and Van Horn, David},
  booktitle={Proceedings of the 2015 ACM SIGPLAN International Conference on Object-Oriented Programming, Systems, Languages, and Applications},
  pages={748--766},
  year={2015},
  doi={10.1145/2814270.2814305}
}

@article{mokhov2018build,
  title={Build Systems \`a la Carte},
  author={Mokhov, Andrey and Mitchell, Neil and Peyton Jones, Simon},
  journal={Proceedings of the ACM on Programming Languages},
  volume={2},
  number={ICFP},
  pages={79:1--79:29},
  year={2018},
  doi={10.1145/3236774}
}

@article{gao2025flowreasoner,
  title={{FlowReasoner}: Reinforcing Query-Level Meta-Agents},
  author={Gao, Hongcheng and Liu, Yue and He, Yufei and Dou, Longxu and Du, Chao and Deng, Zhijie and Hooi, Bryan and Lin, Min and Pang, Tianyu},
  journal={arXiv preprint arXiv:2504.15257},
  year={2025},
  url={https://arxiv.org/abs/2504.15257}
}

@inproceedings{ye2025masgpt,
  title={{MAS-GPT}: Training {LLM}s to Build {LLM}-based Multi-Agent Systems},
  author={Ye, Rui and Tang, Shuo and Ge, Rui and Du, Yaxin and Yin, Zhenfei and Chen, Siheng and Shao, Jing},
  booktitle={Proceedings of the 42nd International Conference on Machine Learning},
  series={Proceedings of Machine Learning Research},
  volume={267},
  pages={72063--72090},
  year={2025},
  url={https://proceedings.mlr.press/v267/ye25g.html}
}

@inproceedings{wu2024autogen,
  title={{AutoGen}: Enabling Next-Gen {LLM} Applications via Multi-Agent Conversations},
  author={Wu, Qingyun and Bansal, Gagan and Zhang, Jieyu and Wu, Yiran and Li, Beibin and Zhu, Erkang and Jiang, Li and Zhang, Xiaoyun and Zhang, Shaokun and Liu, Jiale and Awadallah, Ahmed Hassan and White, Ryen W. and Burger, Doug and Wang, Chi},
  booktitle={First Conference on Language Modeling},
  year={2024},
  url={https://openreview.net/forum?id=BAakY1hNKS}
}

@inproceedings{li2023camel,
  title={{CAMEL}: Communicative Agents for ``Mind'' Exploration of Large Language Model Society},
  author={Li, Guohao and Hammoud, Hasan Abed Al Kader and Itani, Hani and Khizbullin, Dmitrii and Ghanem, Bernard},
  booktitle={Advances in Neural Information Processing Systems},
  volume={36},
  year={2023},
  url={https://arxiv.org/abs/2303.17760}
}

@inproceedings{hong2024metagpt,
  title={{MetaGPT}: Meta Programming for a Multi-Agent Collaborative Framework},
  author={Hong, Sirui and Zhuge, Mingchen and Chen, Jonathan and Zheng, Xiawu and Cheng, Yuheng and Zhang, Ceyao and Wang, Jinlin and Wang, Zili and Yau, Steven Ka Shing and Lin, Zijuan and Zhou, Liyang and Ran, Chenyu and Xiao, Lingfeng and Wu, Chenglin and Schmidhuber, J{\"u}rgen},
  booktitle={International Conference on Learning Representations},
  year={2024},
  url={https://proceedings.iclr.cc/paper_files/paper/2024/file/6507b115562bb0a305f1958ccc87355a-Paper-Conference.pdf}
}

@inproceedings{zhuge2024gptswarm,
  title={{GPTSwarm}: Language Agents as Optimizable Graphs},
  author={Zhuge, Mingchen and Wang, Wenyi and Kirsch, Louis and Faccio, Francesco and Khizbullin, Dmitrii and Schmidhuber, J{\"u}rgen},
  booktitle={Proceedings of the 41st International Conference on Machine Learning},
  series={Proceedings of Machine Learning Research},
  volume={235},
  pages={62743--62767},
  year={2024},
  url={https://proceedings.mlr.press/v235/zhuge24a.html}
}

@inproceedings{liu2024dylan,
  title={A Dynamic {LLM}-Powered Agent Network for Task-Oriented Agent Collaboration},
  author={Liu, Zijun and Zhang, Yanzhe and Li, Peng and Liu, Yang and Yang, Diyi},
  booktitle={First Conference on Language Modeling},
  year={2024},
  url={https://openreview.net/forum?id=XII0Wp1XA9}
}

@inproceedings{zhang2025gdesigner,
  title={{G-Designer}: Architecting Multi-Agent Communication Topologies via Graph Neural Networks},
  author={Zhang, Guibin and Yue, Yanwei and Sun, Xiangguo and Wan, Guancheng and Yu, Miao and Fang, Junfeng and Wang, Kun and Chen, Tianlong and Cheng, Dawei},
  booktitle={Proceedings of the 42nd International Conference on Machine Learning},
  series={Proceedings of Machine Learning Research},
  volume={267},
  pages={76678--76692},
  year={2025},
  url={https://proceedings.mlr.press/v267/zhang25cu.html}
}

@inproceedings{ma2026ann,
  title={Self-Evolving Multi-Agent Systems via Textual Backpropagation},
  author={Ma, Xiaowen and Ma, Yunpu and Lin, Chenyang and Yan, Sikuan and Bi, Jinhe and Cao, Zixuan and Tian, Yijun and Tresp, Volker and Schuetze, Hinrich},
  booktitle={Findings of the Association for Computational Linguistics: ACL 2026},
  pages={9918--9951},
  year={2026},
  doi={10.18653/v1/2026.findings-acl.483},
  url={https://aclanthology.org/2026.findings-acl.483/}
}

@misc{openai2024gpt4omini,
  author={{OpenAI}},
  title={{GPT-4o mini}: Advancing Cost-Efficient Intelligence},
  year={2024},
  howpublished={\href{https://openai.com/index/gpt-4o-mini-advancing-cost-efficient-intelligence/}{OpenAI product announcement}}
}

@misc{openai2026gpt54mini,
  author={{OpenAI}},
  title={Introducing {GPT-5.4 mini} and nano},
  year={2026},
  howpublished={\href{https://openai.com/index/introducing-gpt-5-4-mini-and-nano/}{OpenAI product announcement}}
}

@article{li2026darkforest,
  title={{DarkForest}: Less Talk, Higher Accuracy for Multi-Agent {LLMs}},
  author={Li, Yi and Wei, Songtao and Jiang, Dongming and Guo, Zhichun and Li, Qiannan and Li, Bingzhe},
  journal={arXiv preprint arXiv:2605.25188},
  year={2026},
  url={https://arxiv.org/abs/2605.25188}
}

\clearpage
\appendix
\section{Extended Related Work}
\label{sec:positioning}

\subsection{Workflow Design and Optimization}
Automatic design methods search agent programs~\citep{hu2024adas}, graph prompts and connections~\citep{zhuge2024gptswarm}, or modular agent designs~\citep{shang2024agentsquare}. They differ in both the representation being optimized and the setting in which optimization occurs. In its main benchmark experiments, AFlow~\citep{zhang2024aflow} optimizes candidate workflows on a validation split before evaluating the selected workflow on held-out test data. Its Appendix F also explores single-question optimization with an LLM judge and no reference answers. Query-dependent generation methods such as MaAS~\citep{zhang2025maas}, FlowReasoner~\citep{gao2025flowreasoner}, and MAS-GPT~\citep{ye2025masgpt} further connect workflow construction to individual requests. Inherit-MAS focuses on how an executed task-specific MAS supplies the components and completed computation for its next refinement.

\subsection{Workflow Evolution}
Iterative feedback provides a basis for refinement at several levels. Reflexion~\citep{shinn2023reflexion} carries verbal feedback across trials, and Self-Refine~\citep{madaan2023selfrefine} alternates feedback and response revision. Self-consistency~\citep{wang2023selfconsistency} instead aggregates independently sampled reasoning paths, while test-time compute scaling studies both search and adaptive response refinement~\citep{snell2024testtime}. Inherit-MAS applies feedback to an executable multi-agent workflow and retains earlier candidates for final selection.

Evolving MAS methods carry prior work forward in different forms. EvoAgent~\citep{yuan2024evoagent} executes newly generated agents and integrates their outputs with the previous result. EvoMAS~\citep{hu2026evomas} retains pools of promising configurations and accumulated search experience, mutates one component type that may span several agents, and recombines parent configurations through crossover. TacoMAS~\citep{xu2026tacomas} carries capability state and trajectory-derived feedback across fast and slow loops, with slow updates able to combine changes to the population and communication graph. Retaining such information is distinct from verifying whether an individual stored node result remains valid after a revision. Inherit-MAS makes that decision explicit alongside workflow refinement. Select keeps the incumbent's useful nodes, Edit preserves the kept workflow outside its declared scope, and execution inheritance verifies the complete resolved request and execution context before restoring a result.

Evoflux~\citep{bhandari2026evoflux} is a closely related inference-time method that evolves tool-workflow graphs from execution feedback without updating model weights. It focuses on the execution feasibility of compact planners against live tool servers, whereas Inherit-MAS combines multi-agent workflow refinement with request-verified inheritance of agent outputs.

\subsection{Execution Reuse and Caching}
Execution reuse differs in what is stored, how a match is established, and whether reuse changes the produced result. Demand-driven incremental computation~\citep{hammer2015incremental} and build systems~\citep{mokhov2018build} track dependencies to determine when previous computation remains valid. Inherit-MAS draws on this principle when checking node requests after a workflow edit. At the inference-runtime level, InferCept~\citep{abhyankar2024infercept} addresses the cost of reconstructing prior context around tool interruptions.

Agent systems reuse several kinds of artifacts. StepCache~\citep{nouri2026stepcache} retrieves a similar cached request, verifies the stored response's steps, and regenerates failing regions. Agentic Plan Caching~\citep{zhang2025plancache} adapts plan templates across tasks, and FlowBank~\citep{yuan2026flowbank} precomputes workflows for query-adaptive selection. Helium~\citep{wadlom2026helium} combines cached-operator substitution with common-subgraph elimination. Cost-aware speculative execution~\citep{fareed2026speculative} makes commit barriers and the admissibility of environment-changing actions explicit.

Inherit-MAS combines workflow evolution with task-local inheritance of eligible node outputs, tool observations, and artifacts under a complete resolved-request and execution-context match. Unlike regenerating a cached response's failing regions, an inheritance hit restores the recorded result unchanged, and \Cref{fig:full-eval-cost} measures the execution it avoids against independent reruns of the same controller with inheritance disabled. Whether a fresh model invocation reproduces the stored output is a separate question examined under the deterministic-runtime conditions in Appendix~\ref{sec:store-footprint}.

\section{Additional Ablations and Analysis}

\subsection{Ranking Headroom and Failure Modes}
\label{sec:ranking-headroom}
The online ranking does not merely return the last round: it recovers part, but not all, of the variation created by evolution. On WorkBench, the oracle over stored candidates solves 82 tasks versus 72 returned. On HotpotQA it reaches 56.3\% joint F1 versus 49.7\% returned. This gap motivates better calibrated judges and ensemble ranking, but it also cautions against reporting oracle-any-round as an online system.

The same sealed trajectories (run records that are never modified after a run) also separate initialization, the last generated candidate, and the online ranking without additional model calls. On WorkBench these return policies achieve 49.2\%, 50.0\%, and 55.4\% completion, respectively; on HotpotQA they achieve 44.7\%, 44.4\%, and 49.7\% joint F1. Later candidates are therefore not uniformly better. Refinement creates useful alternatives, while retaining and ranking earlier workflows is important to final performance. The last-candidate view changes only the final return rule; every refinement still builds on the latest candidate.

The stored trajectories provide additional context for the returned-workflow curves in
\Cref{fig:refinement-rounds}. Raw WorkBench candidate completion is not monotonic. It is 46.2, 51.5, 48.5, 42.3, and 50.0\% across candidate indices. The 46.2\% at index 0 counts the six tasks whose initial proposal was invalid as failures, whereas the 49.2\% initialization figure above scores each task's first successfully executed initial workflow, which for those six tasks is the replacement at index 1, four of which succeed. HotpotQA shows the
same nonmonotonic pattern, with joint F1 of 44.7, 46.2, 47.4, 45.9, and
44.4\%. Evolution should therefore be read as dependent iterative search rather
than deterministic hill climbing. Applying the same online ranking rule
to each prefix raises the returned candidate to 55.4\% and 49.7\%, and the
returned candidate's score never decreases on either benchmark. The offline oracle prefixes
rise to 63.1\% and 56.3\%, showing that the main remaining gap is that the judge
does not always identify the best evaluated dependent refinement.

Execution inheritance is concentrated in ordinary refinement rounds. At indices 1 through 3,
the inherited share of worker-token equivalents is between 34.7\% and 44.0\% on WorkBench and between 44.0\% and 63.2\% on HotpotQA. At
index 4 it falls to 17.6\% and 21.0\%, respectively, because the final-round
restart often produces a fresh workflow. The restart's candidate is the returned workflow for 25 WorkBench tasks and 47 HotpotQA questions. The final candidate thus
adds substantial oracle headroom but is also the least inheritance-friendly round.

\subsection{Edit-Level Outcomes and Locality}
\begin{table}[t]
\centering
\caption{Post-hoc incumbent-to-candidate transition aggregates on the sealed trajectories
of the reported Inherit-MAS runs. WorkBench utility is binary completion. HotpotQA utility is joint
F1. Improve is the fraction with strictly positive utility delta. Affected is the mean fraction of nodes in the affected region. Inherit. is the inherited share of worker-token equivalents, and the final column is net $\sum\Delta U$ per million live
worker tokens. Topology covers node insertions, edge edits, and transitions in which Select discarded nodes. These associations are descriptive, not randomized edit effects.}
\label{tab:edit-transitions}
\small
\begin{tabular*}{\textwidth}{@{\extracolsep{\fill}}llrrrrrrr@{}}
\toprule
Benchmark & Edit family & $n$ & Improve & Mean $\Delta U$ & Affected & Inherit. & Live tok. & $\sum\Delta U$/M\\
\midrule
\multirow{5}{*}{WB}
 & Restart & 73 & 9.6\% & $+0.0411$ & 100.0\% & 2.0\% & 4,134 & $+9.94$\\
 & Prompt & 253 & 3.6\% & $+0.0316$ & 70.8\% & 47.3\% & 1,679 & $+18.84$\\
 & Subtask & 78 & 1.3\% & $+0.0000$ & 69.1\% & 47.2\% & 1,568 & $+0.00$\\
 & Tool & 49 & 12.2\% & $-0.0612$ & 100.0\% & 6.3\% & 3,272 & $-18.71$\\
 & Topology & 13 & 7.7\% & $+0.0769$ & 75.0\% & 44.5\% & 2,453 & $+31.36$\\
\midrule
\multirow{5}{*}{HP}
 & Restart & 184 & 19.0\% & $-0.0418$ & 100.0\% & 0.0\% & 4,667 & $-8.95$\\
 & Prompt & 749 & 6.8\% & $-0.0022$ & 52.3\% & 75.0\% & 836 & $-2.68$\\
 & Subtask & 133 & 17.3\% & $+0.0312$ & 91.9\% & 5.6\% & 4,101 & $+7.60$\\
 & Tool & 8 & 25.0\% & $+0.0886$ & 100.0\% & 8.2\% & 3,154 & $+28.10$\\
 & Topology & 106 & 21.7\% & $+0.0344$ & 97.6\% & 4.4\% & 3,625 & $+9.49$\\
\bottomrule
\end{tabular*}
\end{table}

\Cref{tab:edit-transitions} aggregates 466 WorkBench and 1,180 HotpotQA
completed transitions against the incumbent used at proposal time.
Only 24 and 134 transitions, respectively, strictly improve official utility.
Of those, 8/24 (33.3\%) and 37/134 (27.6\%) have an affected region that is a strict subset of the workflow. Thus
most observed improvements are not local under even the permissive definition
that at least one node of the new candidate remains outside the affected region.

Local edits nevertheless provide the clearest cost leverage. Their
token-weighted inheritance is 57.5\% on WorkBench and 81.8\% on HotpotQA, with mean
live worker cost of 1,345 and 596 tokens, compared with 4,134 and 4,667 tokens
for restarts. Quality does not track that locality: local edits
have mean utility deltas of +0.0212 and -0.0007, and the families with positive
means differ across benchmarks (prompt and topology edits and restarts on WorkBench;
subtask, tool, and topology edits on HotpotQA, where restarts have a negative
mean). Prompt edits occupy the useful middle ground, inheriting 47.3\% and 75.0\%
of worker-token equivalents while occasionally improving. The aggregate
evidence therefore supports inheritance as a cost mechanism, but not locality as a
proxy for edit quality. The latter requires a better gain estimator.

Two deterministically selected post-hoc cases make the capability--inheritance
tradeoff concrete. For a
WorkBench reassignment task, adding a context edge from one worker to another
leaves the outcome unchanged, inherits the unaffected worker's prior result, and re-executes
the dependent worker, integrator, and write executor (2,452 inherited worker-token equivalents
versus 760 live worker tokens). For a HotpotQA two-hop question, discarding the
planner and rewriting the researcher's prompt supplies the missing bridge and raises
joint F1 from 0.667 to 1.000, but every node is re-executed (9,614 live worker tokens,
none inherited). Local
rewiring can preserve prior computation. Useful repairs that reshape the workflow may intentionally
invalidate it.

HotpotQA shows a larger win and loss margin (\Cref{tab:paired}), although some of the improvement may come from broader retrieval exposure across candidates rather than from topology changes, which we do not separate here.

\subsection{Paired Headline Effects}
\Cref{tab:paired} reports paired differences and task-level win/loss/tie counts for the main-table comparisons. Wins and losses count strictly better and worse official task scores. Ties include equal scores.

\begin{table}[h]
\centering
\caption{Inherit-MAS minus comparator on each benchmark's primary metric. WB deltas are percentage points. HP deltas are joint F1.}
\label{tab:paired}
\scriptsize
\begin{tabular}{@{}llrr@{}}
\toprule
Bench. & Comparator & $\Delta$ & W/L/T\\
\midrule
WB & Single ReAct (reference) & +13.8 & 29/11/90\\
\paddedhdashline
WB & Initial workflow & +6.2 & 12/4/114\\
WB & EvoAgent & +21.5 & 37/9/84\\
WB & EvoMAS & +24.6 & 39/7/84\\
WB & TacoMAS & +20.8 & 39/12/79\\
\midrule
HP & Single ReAct (reference) & +0.037 & 92/65/143\\
\paddedhdashline
HP & Initial workflow & +0.050 & 54/21/225\\
HP & EvoAgent & +0.149 & 125/32/143\\
HP & EvoMAS & +0.105 & 110/54/136\\
HP & TacoMAS & +0.019 & 81/75/144\\
\bottomrule
\end{tabular}
\end{table}

The HotpotQA TacoMAS row is the adapted implementation described in Appendix~\ref{sec:baseline-fidelity}. The released controller, run through a task and output plugin on the same questions, returned prose answers without the structured supporting-fact array that joint F1 requires, so it has no informative paired joint score and is retained only as a released-interface diagnostic.

\section{Inherit-MAS Implementation}
\label{sec:implementation-appendix}

\subsection{Complete Controller}
\label{sec:controller}
\begin{algorithm}[H]
\caption{Inherit-MAS controller with execution inheritance}
\label{alg:method}
\scriptsize
\algrenewcommand\algorithmicindent{1em}
\begin{algorithmic}[1]
\Require task $x$, task interface $\mathcal I$, cap $K$, pristine environment $e_0$
\State $\mathcal H\gets[\,]$; $\mathcal M\gets\mathsf{EmptyStore}()$
\For{$k=0,\ldots,K-1$}
 \State $p\gets\bot$
 \If{$\mathsf{Complete}(\mathcal H)=\varnothing$}
 \State $G_k\gets\mathsf{ValidatedSynthesis}(x,\mathcal I)$
 \ElsIf{$k=K-1$ and $\mathsf{Restart}(\mathcal H)$}
 \State $G_k\gets\mathsf{ValidatedRestart}(x,\mathcal I,\mathcal H)$ \Comment{fresh restart}
 \Else
 \State $p\gets\mathsf{Latest}(\mathsf{Complete}(\mathcal H))$
 \State $\Delta^{-}\gets\mathsf{Select}(x,G_p,\tau_p,q_p,\gamma_p,\mathcal R(G_p))$;\ \ $G'\gets\mathsf{Prune}(G_p,\Delta^{-})$ \Comment{$\Delta^{-}=\varnothing$ if $\mathcal R(G_p)=\varnothing$}
  \If{$G'$ is invalid}
 \State $\mathcal H.\mathsf{Append}(\mathsf{Invalid}(k))$; \textbf{continue}
 \EndIf
 \State $a_k\gets\mathsf{Propose}(x,G',\tau_p,q_p,\gamma_p,\mathcal A(G'))$
 \State $G_k\gets\mathsf{ValidateApply}(G',a_k,\mathcal I)$
 \EndIf
 \If{$G_k$ is invalid}
 \State $\mathcal H.\mathsf{Append}(\mathsf{Invalid}(k))$; \textbf{continue}
 \EndIf
 \State $D_k\gets V_k$ if $p=\bot$, else $\mathsf{AffectedRegion}(G_p,G_k)$
 \State $e_k\gets\mathsf{Fork}(e_0)$;\ \ $\phi_k\gets\mathsf{Fingerprint}(e_k)$
 \For{$v$ in a topological order of $G_k$}
 \State $r_{k,v}\gets\mathsf{Resolve}(x,v,[(\pi_i,w_{k,u_i})]_{i=1}^{d_v},\phi_k)$
  \State $\kappa_{k,v}\gets h(r_{k,v})$
 \If{$\mathsf{Eligible}(v,e_k)$ and $v\notin D_k$ and $G_p$ has an eligible stored record for $v$}
 \State $(o,A,t,\epsilon)_{k,v}\gets\mathsf{VerifyIncumbentRecord}(G_p,v,\kappa_{k,v})$
 \ElsIf{$\mathsf{Eligible}(v,e_k)$ and $\mathsf{ValidRecord}(\mathcal M,\kappa_{k,v})$}
 \State $(o,A,t,\epsilon)_{k,v}\gets\mathcal M[\kappa_{k,v}]$
 \Else
 \State $(o,A,t,\epsilon,\chi)_{k,v}\gets\mathsf{Execute}(v,r_{k,v},e_k)$
 \If{$\chi_{k,v}$}
 \State $\mathsf{Store}(\mathcal M,\kappa_{k,v},(o,A,t,\epsilon)_{k,v})$
 \EndIf
 \EndIf
 \State $w_{k,v}\gets\mathsf{Payload}(v,o_{k,v},A_{k,v},\epsilon_{k,v})$
 \EndFor
 \State $(y_k,\tau_k,c_k)\gets\mathsf{Assemble}(G_k,\{(o,A,\epsilon)_{k,v}\}_{v\in V_k},e_k)$
 \State $b_k\gets\mathsf{HardFailures}(y_k,\tau_k)$
 \State $(q_k,\gamma_k)\gets\mathsf{Judge}(x,G_k,y_k,\tau_k)$
 \If{judgment remains invalid after the permitted repair}
 \State $\mathcal H.\mathsf{Append}(\mathsf{Invalid}(k))$; \textbf{continue}
 \EndIf
 \State $s_k\gets\mathsf{RankingKey}(y_k,q_k,b_k,k)$
 \State $\mathcal H.\mathsf{Append}(k,G_k,y_k,\tau_k,q_k,\gamma_k,s_k,c_k)$
\EndFor
\If{$\mathsf{Complete}(\mathcal H)=\varnothing$}
\State \Return failure
\EndIf
\State \Return $y_{\arg\max_{i\in\mathsf{Complete}(\mathcal H)}^{\mathrm{prio}}s_i}$
\end{algorithmic}
\end{algorithm}

Here $p=\bot$ denotes a candidate synthesized without a parent, and $\epsilon_{k,v}$ is the execution-error record exposed to assembly and judging; an inherited record restores its stored $\epsilon$, so a stored error still counts toward $b_k$. $\mathsf{VerifyIncumbentRecord}$
checks the incumbent record against $\kappa_{k,v}$, so a mismatch is a blocking
integrity error rather than a silent miss. The flag $\chi_{k,v}\in\{0,1\}$ indicates whether the result satisfies the
node's eligibility contract and may enter $\mathcal M$.
The write-capable sink is never eligible for either inheritance branch. A read-only
node may contain a bounded multi-step tool loop. Its pre-execution request includes
the rendered node instructions, ordered upstream payloads, effective execution limits,
and execution-context fingerprint. Observations produced inside that loop are stored
with the result and restored on an inheritance hit; they are not part of the
pre-execution lookup key.
$\mathsf{Complete}(\mathcal H)$ is the set of indices of candidates with recorded execution and a valid judgment; $\mathsf{Latest}$ returns the most recent such index. $\mathcal R(G_p)$ lists the nodes whose individual removal leaves the workflow valid; $\mathsf{Select}$ is validated like $\mathsf{Propose}$, so a discard set outside $\mathcal R(G_p)$ or a kept workflow that fails validation is rejected, one JSON repair is permitted, and a second failure records an invalid proposal. $\mathsf{Restart}(\mathcal H)$ holds when $\mathsf{Complete}(\mathcal H)$ is empty, when no candidate in it has a valid output, or when $\max_{i\in\mathsf{Complete}(\mathcal H)} q_i^{\mathrm{all}}\le q_{i_0}^{\mathrm{all}}+1$, where $i_0$ is the first completed candidate. $\mathsf{ValidatedRestart}$ supplies the highest-ranked candidate's judge feedback and workflow to the restart prompt (Appendix~\ref{sec:prompts}).
The restart is recorded as a full-graph transition with no workflow inheritance. It fires on 73/466 completed WorkBench transitions
(15.7\%) and 184/1,180 HotpotQA transitions (15.6\%).

Throughout the paper, \emph{initial workflow} means the first successfully executed candidate produced by initial synthesis. If synthesis fails before that point, the failed attempts remain in proposal-validity and cost accounting but are not treated as executable workflows.

\subsection{Inherit-MAS Settings}
\Cref{tab:implementation} lists the settings shared by the reported Inherit-MAS runs on both benchmarks. Benchmark-specific roles, tools, output schemas, and runtime limits are given in \Cref{sec:workflow-rules,sec:runtime-limits}.
\begin{table}[h]
\centering
\caption{Headline Inherit-MAS implementation settings shared across benchmarks unless noted.}
\label{tab:implementation}
\scriptsize
\begin{tabular}{@{}p{0.34\linewidth}p{0.61\linewidth}@{}}
\toprule
Component & Frozen setting\\
\midrule
Workers & GPT-4o-mini, temperature 0\\
Synthesis, node selection, refinement, judge & GPT-5.4-mini, separate judge prompt, temperature 0\\
Search & 5 candidates, 1 Select decision and 1 validated edit per refinement, no early stop\\
Structured-output recovery & At most 1 JSON-repair call\\
Provider recovery & At most 1 identical retry for HTTP 500\\
HotpotQA retrieval & BM25, top 5, at most 4 calls/candidate\\
Inheritance scope & Task-local, request-exact, read-only nodes\\
Stateful commit & One write-capable sink, always executed live\\
API/runtime & Azure OpenAI-compatible endpoint\\
\bottomrule
\end{tabular}
\end{table}

\subsection{Workflow and Output Rules}
\label{sec:workflow-rules}
\paragraph{Edit family and validation.}
The workflow specification declares each node's role, accepted inputs, produced output, and tool permissions; every edge must carry an allowed output to a compatible, ordered input. These explicit rules let the proposer choose from a finite menu of edits whose arguments can be checked before execution. An edit may change a node's prompt, assigned subtask, tool access, communication, or topology while still identifying where its influence begins. The controller is shared across benchmarks. Each adapter supplies the allowed node roles, input and output labels, input ordering, tools, output validator, and execution constraints through $\mathcal I$. The implemented edit family is
\begin{align}
\mathcal A(G)\subseteq\{&
\mathsf{Prompt}(v),\mathsf{Subtask}(v),\mathsf{Tools}(v),\nonumber\\
&\mathsf{AddNode}(v),\mathsf{InsertNode}(u,v),\mathsf{DropNode}(v),\nonumber\\
&\mathsf{AddEdge}(u,v),\mathsf{DropEdge}(u,v),
\mathsf{Reorder}(v)\}.
\end{align}
$\mathsf{InsertNode}(u,v)$ places a new node on an existing edge $u\to v$: the new node takes $u$'s output and feeds $v$, and the edge $u\to v$ is retained. Only arguments permitted by these role, connection, and tool rules are instantiated in $\mathcal A(G)$. The validator enforces acyclicity, input/output compatibility and input cardinalities, identifier uniqueness, output reachability, authorized tools, and benchmark execution constraints.

\paragraph{HotpotQA.}
The synthesis payload contains this graph shape:
\begin{lstlisting}
{
 "nodes": [
 {"id":"researcher_a", "role":"researcher",
 "subtask":"...", "system_prompt":"...",
 "tools":["search_fullwiki"], "inputs":[]},
 {"id":"answerer", "role":"synthesizer",
 "subtask":"...", "system_prompt":"...", "tools":[],
 "inputs":[{"from":"researcher_a","port":"evidence"}]}
 ],
 "output":"answerer"
}
\end{lstlisting}
A \texttt{planner} emits \texttt{plan} and has no inputs. A
\texttt{researcher} emits \texttt{evidence} and may consume plans or evidence.
A \texttt{verifier} emits \texttt{critique} and may consume evidence or
critique. A \texttt{synthesizer} emits \texttt{answer} and may consume plans,
evidence, or critique. Every edge port must equal its source role's emitted
type. Only planners and researchers may be roots. The graph has two to seven
nodes, exactly one synthesizer, at least one retrieval-capable researcher, and
one synthesizer output that is the only sink.

\paragraph{WorkBench.}
The synthesis payload contains this graph shape:
\begin{lstlisting}
{
 "name":"short name",
 "nodes":[
 {"id":"worker_a", "role":"worker", "subtask":"...",
 "system_prompt":"...", "tools":["declared read-only tool"],
 "inputs":[]},
 {"id":"integrator", "role":"integrator",
 "subtask":"combine evidence", "system_prompt":"...",
 "tools":[], "inputs":[{"port":"evidence_a","from":"worker_a"}]},
 {"id":"executor", "role":"executor",
 "subtask":"execute approved plan", "system_prompt":"",
 "tools":[], "inputs":[{"port":"plan","from":"integrator"}]}
 ],
 "sink":"executor",
 "rationale":"brief"
}
\end{lstlisting}
The allowed roles are \texttt{worker}, \texttt{critic},
\texttt{integrator}, \texttt{verifier}, and \texttt{executor}. Only workers
may be roots or call read-only tools. Critics consume worker, critic, or
integrator reports. Integrators consume worker or critic reports. A verifier
receives exactly one plan from an integrator and may also receive worker or
critic evidence. The graph contains at most eight language-model nodes and
twelve ordered edges. Exactly one deterministic executor is the sole sink. It
receives one plan from an integrator or verifier, has no prompt or tools, and
is the only node allowed to execute state-changing calls.

\subsection{Prompts, Scores, and Role Guides}
\label{sec:prompts}
This section reproduces the prompt templates used by the reported Inherit-MAS
runs on both benchmarks. They are copied verbatim from the implementation. Line wrapping and
quotation marks are typographic only. Task text, graph-specific instructions,
upstream records, and tool schemas are substituted at runtime and serialized
without paraphrasing. Every synthesized \texttt{subtask} and
\texttt{system\_prompt} is stored in the trajectory record.

\paragraph{Model roles.}
GPT-5.4-mini serves five controller roles: workflow synthesizer,
node selector, edit proposer, structured judge, and one-shot JSON repair model.
GPT-4o-mini executes every language-model node in the synthesized workflow.
The WorkBench write executor and all schema validators are deterministic.
HotpotQA retrieval uses the fixed BM25 service. Thus the judge and proposer use
different system messages even though they use the same model family.

\paragraph{Initial workflow synthesis.}
The controller system message is:
\begin{quote}\small
``Synthesize a small typed multi-agent workflow for the supplied task and
interface. Use only the declared roles, ports, and tools. Decompose only when
useful, keep information flow explicit, and end at the declared output node.
Return JSON only.''
\end{quote}
Here, \emph{typed} is the wording of the frozen prompt and means only that node roles, input/output labels, and tool permissions must follow the explicit workflow rules reproduced below.
The user message is a JSON object with fields \texttt{benchmark},
\texttt{task}, \texttt{interface\_contract}, and
\texttt{workflow\_contract}. The final-round restart appends
``This is a stagnation escape; produce a meaningfully different valid
workflow.'' Its payload adds \texttt{reason}, \texttt{prior\_audit}, and
\texttt{prior\_graph}. We treat this restart as an implementation detail rather than
a separate method component.

\paragraph{Structured judge.}
The judge system message is:
\begin{quote}\small
``You are an independent, skeptical, gold-free evaluator of a task workflow.
Derive the task's requirements from the public request and interface contract.
Use only the proposed output, the observed artifacts in the audit, and the
workflow's recorded node outputs. Penalize unsupported claims, missing
requirements, invalid actions, and fluent guessing. Do not assume any
benchmark-specific number of sources, steps, agents, or actions. When the node
outputs show which workflow nodes contributed useful or useless work, name
those node ids in the critique and in the keep/fix strings. Scores are 0--100.
Return compact JSON with exactly the contract's fields and no others.''
\end{quote}
The user message contains \texttt{benchmark}, \texttt{task}, the public
\texttt{interface\_contract}, \texttt{proposed\_output},
\texttt{artifact\_audit}, the candidate's \texttt{workflow}, and its
\texttt{node\_outputs} (the compact execution trace), followed by this exact
response contract:
\begin{lstlisting}
{
 "quality_score": 0,
 "completion_likelihood": 0,
 "requirement_coverage": 0,
 "artifact_grounding": 0,
 "critique": "brief skeptical audit",
 "keep": ["component worth preserving"],
 "fix": ["highest-value repair"]
}
\end{lstlisting}
All four numeric fields must lie in $[0,100]$. The critique must be nonempty.
The validator retains at most four nonempty strings from each of
\texttt{keep} and \texttt{fix}. Search cost is not part of
\texttt{quality\_score}. The online ranking first minimizes hard execution
failures, then prefers a valid parsed output, \texttt{quality\_score},
\texttt{requirement\_coverage}, and \texttt{artifact\_grounding}, in that
order. The earlier candidate wins an exact tie.

The \texttt{artifact\_audit} supplied to the judge has this common shape:
\begin{lstlisting}
{
 "output_valid":true,
 "artifacts":[
 {"artifact_id":"...", "source":"...",
 "locator":"...", "observed":true,
 "content":"bounded observed content"}
 ],
 "execution_errors":{"node_id":"error"},
 "interface_diagnostics":{...}
}
\end{lstlisting}
For WorkBench, artifacts are read-only tool observations and interface
diagnostics contain the proposed actions and invalid-output count. For
HotpotQA, each artifact is a cited title and sentence index, and diagnostics
contain declared and observed citation counts. The judge receives bounded
observed content rather than a benchmark reference answer.

\paragraph{Node selection.}
The node-selection system message is:
\begin{quote}\small
``Select which nodes of the latest workflow the next candidate inherits. Use
the gold-free audit and execution diagnostics. Discard a node only when the
audit or its recorded output shows that it contributes nothing useful or harms
the result, and keep every other node. The kept workflow must remain valid
under the workflow contract. Return compact JSON only.''
\end{quote}
Its JSON user message contains \texttt{benchmark}, \texttt{task},
\texttt{interface\_contract}, \texttt{workflow\_contract}, the incumbent
\texttt{graph}, the judge \texttt{audit}, \texttt{artifact\_audit}, a compact
\texttt{execution} trace, \texttt{discardable\_nodes}, and the following
response contract:
\begin{lstlisting}
{
 "discard": ["id of a node to drop; leave the list empty to keep every node"],
 "rationale": "why the kept nodes should be inherited"
}
\end{lstlisting}
\texttt{discardable\_nodes} lists the nodes whose individual removal leaves
the workflow valid, so the output node and any node that another node requires
never appear in it. A discard set naming any other node, or one whose kept
workflow fails validation, is rejected. When the list is empty, the call is
skipped and the incumbent is inherited whole.

\paragraph{Edit proposal.}
The proposer system message is:
\begin{quote}\small
``Choose exactly one available, prevalidated atomic workflow edit. Use the
gold-free audit and execution diagnostics, preserve useful components, and
address the highest-value reported defect. Do not emit a compound edit or copy
a current value unchanged. Return compact JSON only.''
\end{quote}
Its JSON user message contains \texttt{benchmark}, \texttt{task},
\texttt{interface\_contract}, the kept workflow as \texttt{graph}, the
\texttt{selection} (discarded node ids and rationale), the judge
\texttt{audit}, \texttt{artifact\_audit}, a compact
\texttt{execution} trace, an indexed \texttt{available\_edits} menu, and the
following response contract:
\begin{lstlisting}
{
 "edit_index": 0,
 "rationale": "why this one atomic edit addresses the audit",
 "replacement": "only when required by the menu entry",
 "subtask": "only for an add-node edit that requires it",
 "system_prompt": "only for an add-node edit that requires it"
}
\end{lstlisting}
Previously tried indices for the same kept workflow are removed before prompting. Text
edits also expose their current value and require a different replacement.
The benchmark adapter enumerates only edits whose result passes graph validation. The
WorkBench menu includes prompt and subtask replacement, worker-tool addition or
removal, node removal, input reordering, edge addition or removal, and worker
insertion on an existing edge. The HotpotQA menu includes prompt and subtask
replacement, retrieval toggling, edge addition or removal, node removal,
researcher addition, and researcher insertion on an existing edge.

\paragraph{Structured-output repair.}
The controller permits one JSON-repair attempt after a parsing or validation
failure. Its system message is ``Repair one JSON response. Address the exact
validator error. Return JSON only.'' The JSON user message contains
\texttt{validator\_error}, \texttt{contract},
\texttt{invalid\_response}, \texttt{original\_system}, and
\texttt{original\_user}. A second failure remains an invalid proposal or
judgment and is not silently resampled.

The compact execution trace contains aggregate live and inherited token equivalents, tool-call
count, and one record per node. Each node record includes role, execution error,
inheritance status, and a bounded output excerpt. WorkBench additionally includes
active domains, parser failures, rejected artifact references, and the observed
artifact count. HotpotQA includes the number of BM25 calls. This is the complete
execution-diagnostic surface available to the edit proposer.

\subsection{Worker Role Prompts}
The following wrappers guide every GPT-4o-mini node. A synthesized node prompt
is inserted at \texttt{NODE\_INSTRUCTION}. Because this value is part of the
graph, prompt edits change the resolved request and invalidate the corresponding
stored record.

\paragraph{WorkBench shared wrapper.}
Every language-model node receives:
\begin{quote}\small
``You are one read-only node in a multi-agent workflow. Follow your assigned
subtask. Do not claim access to hidden gold outcomes. Do not execute
state-changing actions.

\texttt{NODE\_INSTRUCTION}''
\end{quote}
The role-specific suffix is one of the following:
\begin{itemize}[leftmargin=*]
 \item \textbf{Worker:} ``Use assigned read-only tools when needed. Preserve
 concrete IDs, values, and complete findings in your final response.''
 \item \textbf{Integrator:} ``Produce only JSON in the following form. Use
 the fewest state-changing actions that fully satisfy the task. Exact schemas:
 \texttt{WRITE\_TOOL\_SCHEMAS}.''
\begin{lstlisting}
{"actions":[{"tool":"...","args":{...}}]}
\end{lstlisting}
 \item \textbf{Verifier:} ``Audit the proposed plan for task coverage,
 grounded arguments, and harmful extras. Return only JSON in the following
 form. \texttt{WRITE\_TOOL\_SCHEMAS}.''
\begin{lstlisting}
{"verdict":"approve|revise",
 "actions":[...],"reasons":[...]}
\end{lstlisting}
 \item \textbf{Critic:} ``Return a concise critique or reasoning artifact for
 downstream nodes.''
\end{itemize}
The user-message template is:
\begin{lstlisting}
CURRENT_DATETIME
Task: TASK_TEXT
Assigned subtask: NODE_SUBTASK
Upstream records: ORDERED_JSON_RECORDS
\end{lstlisting}
Each upstream record contains its input port, node identifier, role, text
output, tool events, and error. The prompt expands
\texttt{WRITE\_TOOL\_SCHEMAS} to the exact fourteen official WorkBench write
signatures and their enumerated argument constraints. The synthesis interface
also includes the exact upstream documentation for the thirteen read-only
tools. These benchmark-owned schemas are inserted verbatim, including
qualified tool names, parameter types, descriptions, and examples. The
recorded source lock contains the complete expansion rather than a shortened
rewording.

\paragraph{HotpotQA shared wrapper.}
Every node receives:
\begin{quote}\small
``You are one node in a HotpotQA FullWiki multi-agent DAG. The retrieval tool
is deterministic BM25. Use only retrieved observations and ordered upstream
reports. Retrieved sentences are addressed by exact [title, sentence\_id]
pairs. Never invent a title, sentence index, or fact.''
\end{quote}
The wrapper appends exactly one role instruction:
\begin{itemize}[leftmargin=*]
 \item \textbf{Planner:} ``Decompose the two-hop question into precise
 retrieval subgoals.''
 \item \textbf{Researcher:} ``Retrieve and report concise bridge/comparison
 evidence. Include exact [title, sentence\_id] citations for every useful
 fact and distinguish evidence from hypotheses.''
 \item \textbf{Verifier:} ``Audit upstream evidence, resolve contradictions,
 and search only if necessary. Return the supported answer candidate and
 exact [title, sentence\_id] citations.''
 \item \textbf{Synthesizer:} ``Return JSON only in the following form. Cite
 only observed indexed sentences.''
\begin{lstlisting}
{"answer":"short answer",
 "supporting_facts":[["exact title",0]],
 "confidence":0.0,"rationale":"brief"}
\end{lstlisting}
\end{itemize}
It then appends ``Node instruction:'' and the synthesized node system prompt.
The user-message template is:
\begin{lstlisting}
Question:
QUESTION_TEXT

Assigned subtask:
NODE_SUBTASK

Ordered upstream reports:
ORDERED_JSON_REPORTS
\end{lstlisting}
An upstream report contains its declared input port, source role, text output, and error.

\subsection{Output Schemas and Runtime Limits}
\label{sec:runtime-limits}
The WorkBench integrator returns an \texttt{actions} list whose entries contain
a string \texttt{tool} and an object \texttt{args}. A verifier returns
\texttt{verdict}, \texttt{actions}, and \texttt{reasons}. Only actions that
pass the official tool-name and argument-schema validator reach the
deterministic executor. Worker-role nodes run for at most eight turns, 1,024 output
tokens per turn, and twenty read-only tool calls. Integrator, verifier, and
critic calls have a 1,800-token output limit.

The HotpotQA synthesizer returns a string \texttt{answer}, a list of
\texttt{supporting\_facts}, numeric \texttt{confidence}, and string
\texttt{rationale}. Every supporting fact is a title and integer sentence-index
pair. A citation is valid only if the exact pair appeared in this candidate's
retrieval observations. Each candidate receives four BM25 calls in total, allocated deterministically across retrieval-capable nodes, so adding or removing such a node changes the allowance of the others; a node whose allowance changes joins the affected region of \Cref{eq:affected-region}. Each call returns five documents. A tool-enabled node runs for at most five turns and all language-model nodes have a 900-token output limit.

For meta-model and judge calls, WorkBench uses output limits of 5,000 tokens for initial
synthesis, 5,000 for refinement, and 3,000 for judging. HotpotQA uses 1,600,
900, and 500 tokens, respectively. Worker decoding uses temperature zero. The
full resolved messages, model identity, decoding parameters, tool contract,
execution-context fingerprint, and ordered inputs are recorded with every stored
record.

\subsection{Search Inputs and Offline Scoring}
\label{sec:search-inputs}
For WorkBench, the meta-model and judge receive the natural-language request, public tool names and schemas, tool observations produced by their own candidate, messages, parser errors, and timeouts. For HotpotQA, they receive the question and results of their own bounded BM25 calls. Reference actions, answers, supporting facts, and official metrics are reserved for post-run scoring. 

The judge uses a system prompt separate from the synthesis/refinement prompt and additionally receives the candidate's workflow and each node's recorded output, which lets its critique name nodes. Its structured output contains four scores in $[0,100]$: overall quality, completion likelihood, requirement coverage, and artifact grounding. The online ranking first rejects hard execution failures, then prefers valid output, overall quality, requirement coverage, artifact grounding, and finally the earlier candidate as a deterministic tie-break.

\section{Evaluation Protocol and Baselines}
\label{sec:reproducibility}
The WorkBench manifest contains 130 tasks, all tasks from 13 held-out template clusters. The HotpotQA manifest contains 300 public IDs selected by a deterministic hash and excludes all earlier development IDs. The WorkBench Overall column is the completion rate over all evaluation tasks, so domains contribute in proportion to their task counts: 20 each for Analytics, Calendar, Email, and Project management, 10 for CRM, and 40 for Multi-domain. HotpotQA joint F1 follows the benchmark evaluator: for each question, joint precision and recall are the products of the answer and supporting-fact precision and recall, joint F1 is their harmonic mean, and the table reports the mean over questions. Joint F1 therefore never exceeds the smaller of answer F1 and supporting-fact F1 and penalizes a correct answer without valid citations. Stored-record integrity audits pass for 1,859 inherited WorkBench results over 230 development tasks and 751 inherited HotpotQA results over 200 development questions.

\subsection{External Baselines}
\label{sec:baseline-fidelity}
The baseline implementations use pinned upstream releases with benchmark-specific adaptations. We retain released controllers where supported and otherwise adapt their search mechanisms to the benchmark interface. This subsection describes the GPT-4o-mini configurations; Appendix~\ref{sec:qwen-settings} gives the Qwen configurations. The adaptations and their source/configuration provenance are documented below.

\paragraph{Model assignments and decoding.}
EvoAgent uses GPT-4o-mini across roles. EvoMAS and TacoMAS assign GPT-4o-mini to designated worker and compiler calls and GPT-5.4-mini to meta-model and judge calls, with EvoMAS's worker palette fixed to GPT-4o-mini. TacoMAS's additional answer-synthesis routing is specified in Appendix~\ref{sec:qwen-settings}. Native meta-model calls use temperature 0.7 for EvoMAS and 0.3 for TacoMAS; the HotpotQA TacoMAS adaptation uses temperature zero. Workers and judges use temperature zero, except that native EvoMAS workers follow the decoding settings in each evolved configuration.

\paragraph{EvoAgent.}
We port the population mechanism from commit \texttt{fc6d087} to the benchmark interfaces. The WorkBench configuration follows the released interactive path without a quality-filter call; HotpotQA retains role-quality filtering and permits four BM25 calls per agent. An auxiliary mixed-model configuration uses GPT-5.4-mini for role generation and quality filtering with GPT-4o-mini workers. It reaches 27.7\% WorkBench completion and 30.2\% HotpotQA joint F1, with 73.2\% answer F1 and 38.4\% supporting-fact F1. The HotpotQA comparison isolates model assignment. The WorkBench comparison also changes the quality-filter step, so its difference cannot be attributed to the model alone. We report the all-GPT-4o-mini configuration in the main tables.

\paragraph{WorkBench EvoMAS.}
We executed EvoMAS at commit \texttt{93fd9d6} as one serial dependent trajectory over the WorkBench manifest. The run retains cross-task memory, persistent per-domain pools, native mutation/crossover, model mutation, reward computation, experience consolidation, pool admission, parent rechecks, and fresh final execution. Its declared hyperparameters are matched to our search allowance: two parents, three dependent evolution rounds yielding five unique candidates, CodeAgent \texttt{max\_steps=8}, and a 1,024-token per-request output cap. The model-mutation operator is retained, but the worker palette is fixed to GPT-4o-mini as on HotpotQA, and a post-run trace audit confirms that all 168.5M worker tokens executed on GPT-4o-mini. The fresh reevaluation of the selected winner completes 40 tasks (30.8\%) using 173.59M provider tokens and \$36.00. This is an algorithm-faithful hyperparameter match with matched role-level models, not a released-default reproduction.

\paragraph{HotpotQA EvoMAS.}
The GPT-4o-mini row uses the released controller at commit \texttt{93fd9d6} with a public task and output plugin. It retains serial task order, persistent pools and memory, two parents, two evolution steps per task, and model mutation within the GPT-4o-mini worker palette. Each native agent receives four calls to the pinned BM25 service. The controller executes independently of our graph representation and edit system.

The model-matched EvoMAS run completed one continuous state chain over all HotpotQA questions. A trace-level audit over 2,100 trajectory files and 10,535 worker traces found only GPT-4o-mini workers. A separate scoring amendment parses exact Python dictionary serialization without inferring content from prose. The source lock, trace audit, and sealed official score are retained in the experiment records.

\paragraph{HotpotQA TacoMAS.}
The main-table implementation adapts the fast capability and slow topology mechanisms from commit \texttt{6f0d545} to our shared execution runtime and structured answer/supporting-fact interface. It uses ten fast rounds, birth/death and rewiring checks every two rounds, five to twenty agents, at most two birth/death pairs and eight edge edits per update, and temperature zero. Each candidate shares four BM25 calls across its retrieval nodes, with the same sentence-mapping contract and no execution inheritance. This run uses GPT-4o-mini workers and GPT-5.4-mini meta-models and judges on the same manifest as the Inherit-MAS run. Qwen TacoMAS uses the same adaptation.

We also ran the released TacoMAS controller with a task and output plugin, preserving its native population, graph, ten fast rounds, and slow topology updates. It returned prose answers: 126 contained no citation and only 58 mentioned a sentence index. Its 33.2\% answer F1 is retained as a released-interface diagnostic; missing structured supporting facts prevent an informative joint F1 comparison.

\paragraph{WorkBench TacoMAS.}
We executed TacoMAS at commit \texttt{6f0d545} on all main-table WorkBench tasks with its native AgentState, GraphManager, EvolutionController, five-agent initial population, birth/death, rewiring, ten fast rounds, and slow updates. Because the release does not implement WorkBench's canonical stateful executor, the benchmark bridge gives evolving agents thirteen canonical read-only tools and a non-mutating \texttt{done}. One post-evolution GPT-4o-mini compiler maps the final plan to official write schemas, and one deterministic executor applies the writes to a pristine sandbox. No TacoMAS state is projected into our graph representation. The run completes 45 tasks (34.6\%) using 236.59M provider tokens and \$50.72.

This is a benchmark adaptation of the released controller, not a verbatim upstream WorkBench implementation. We retain the older released-interface run, which solved 14/115 tasks, all of them no-action tasks, at 22.11M tokens and \$12.10, only as an interoperability diagnostic. Unlike that run, the main-table adaptation is scored over the identical manifest with official execute-then-state-difference semantics.

\begin{table}[h]
\centering
\caption{Descriptive released-interface TacoMAS diagnostic, retained separately from the main-table adaptation.}
\label{tab:taco-workbench-interface}
\scriptsize
\begin{tabular}{@{}lrrrr@{}}
\toprule
System & Tasks & Complete & Total live & USD\\
\midrule
TacoMAS & 115 & 14 (12.2\%) & 22.11M & 12.10\\
\bottomrule
\end{tabular}
\end{table}

The WorkBench EvoMAS run completes a continuous state chain over all WorkBench tasks, 390 dependent evolution rounds, 37,207 successful provider calls, 173.59M tokens, and \$36.00 of estimated API spend. The model-matched HotpotQA EvoMAS run completes a continuous state chain over all HotpotQA questions, 600 evolutionary operations, 34,157 successful model calls, 112.28M tokens, and \$29.74. Every observed worker trace in both runs uses GPT-4o-mini. The released-interface HotpotQA TacoMAS diagnostic run completes 2,992 fast rounds and 1,362 slow updates with 75,156 successful calls, 202.68M tokens, and \$56.65. Two failed provider calls remain within otherwise completed trajectories. The WorkBench TacoMAS run completes 1,290 fast rounds and 578 slow updates with 39,699 successful native calls, 236.59M tokens, and \$50.72. Its one oversized final plan is retained as an incorrect compiler outcome. All final outcomes remain in each official denominator.

\subsection{Qwen3-32B Backbone Evaluation}
\label{sec:qwen-settings}
The completed Qwen evaluation uses the same WorkBench and HotpotQA tasks across all five systems. Inherit-MAS replaces only its worker model, retaining its synthesis/refinement controller, judge, prompts, edit family, and scoring rules. Workers use Qwen3-32B in non-thinking mode at temperature zero, while synthesis/refinement and judging remain assigned to GPT-5.4-mini. EvoAgent, which has no separate meta-model, instead uses Qwen3-32B for every logical role, including role generation, integration, and HotpotQA role-quality filtering, through the same benchmark bridge as its GPT-4o-mini row and with no API calls. Serving uses vLLM 0.21.0 with bfloat16 weights, tensor parallelism two, a 32,768-token context, at most two concurrent sequences per replica, batch-invariant execution, CUDA graphs, and disabled automatic prefix caching. Model and tokenizer metadata and serving flags are recorded with the runs. Scheduling-only acceleration does not change these inference settings.

For WorkBench, the Qwen EvoMAS run uses its released evolution loop with two initial parents, three evolution rounds, five unique candidates, eight CodeAgent steps, and a 1,024-token worker completion limit. It preserves serial cross-task experience. The Qwen TacoMAS run retains the released fast/slow evolution mechanism and benchmark executor adapter. Task-solving calls, including worker-facing memory/top-$k$ synthesis and final action compilation, are constrained to Qwen rather than allowed to fall back to the meta-model. In contrast, the historical GPT-4o-mini TacoMAS configuration retains native meta-model routing for memory/top-$k$ answer synthesis. Thus, its cross-backbone comparison also includes this explicit role-routing correction. Native meta-model decoding settings are retained, including TacoMAS's temperature 0.3. These role constraints and per-method search settings are part of the reported implementation.

On HotpotQA, the Qwen EvoMAS and TacoMAS implementations share the execution runtime and structured answer/supporting-fact interface described in Appendix~\ref{sec:baseline-fidelity}. Qwen EvoMAS uses two parents, two evolution steps, four candidate attempts, and per-question population initialization without persistent cross-task experience; its GPT-4o-mini counterpart retains the released controller. Qwen TacoMAS uses the same graph adaptation as its GPT-4o-mini counterpart. Both Qwen adaptations use temperature zero, four BM25 calls per candidate, the same sentence-mapping contract, and no execution inheritance. Cross-backbone comparisons therefore include the stated implementation and model-routing differences.

Failed tasks remain in the denominator. Only completed, sealed runs with working retrieval infrastructure are scored. Local Qwen tokens and Azure meta-model and judge tokens are recorded separately. A local model's absence of an API charge does not imply zero inference cost, and raw token counts across different tokenizers are not treated as equal compute.

\section{Cost Accounting}
\label{sec:cost-appendix}

\subsection{Inference and Search Cost}
\begin{figure}[t]
\centering
\includegraphics[width=0.94\textwidth]{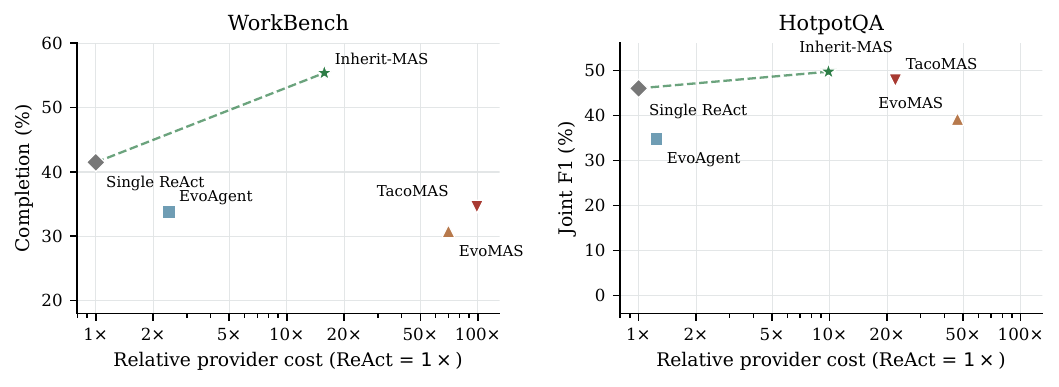}
\caption{Observed quality--cost operating points on the complete benchmark manifests. Provider cost is normalized within each benchmark to Single ReAct $=1\times$ using the configured deployment tariffs; the logarithmic horizontal axis favors lower cost and the vertical axis favors higher quality. The dashed segment joins the observed non-dominated points and is not an interpolation. Inherit-MAS is more accurate and less costly than EvoMAS and TacoMAS under these configurations, while Single ReAct remains a cheaper non-MAS reference.}
\label{fig:cost-performance}
\end{figure}

In \Cref{tab:inference-cost}, \emph{Worker} covers task-solving calls, including integration and benchmark-output compilation, while \emph{Meta/judge} covers workflow synthesis, node selection, refinement, and critique. This is a role-based accounting split: EvoAgent uses GPT-4o-mini for its internal meta-model calls as well as its task-solving calls. \Cref{fig:cost-performance} reports the observed quality--cost positions, while \Cref{tab:inference-cost} provides the provider ledger, and \Cref{tab:cost,tab:ours-cost-ledger,tab:prefix-cost} add inherited equivalents, wall time, and candidate-prefix costs. The role split exposes a different observed allocation of inference. Worker calls account for 13.9\% of Inherit-MAS provider tokens on WorkBench, compared with 97.1\% for EvoMAS and 94.0\% for TacoMAS. On HotpotQA, the corresponding shares are 43.9\%, 92.5\%, and 51.8\%. The absolute totals remain essential: EvoMAS and TacoMAS respectively use 16.8$\times$/22.9$\times$ the Inherit-MAS WorkBench total and 13.7$\times$/2.1$\times$ its HotpotQA total. Execution inheritance avoids 0.53M WorkBench and 1.93M HotpotQA worker-token equivalents, while meta-model and judge inference remains live. Under these configurations, Inherit-MAS combines stronger task-conditioned workflows, workflow inheritance, and execution inheritance to achieve a better observed quality--cost operating point than EvoMAS and TacoMAS. These provider-token counts are not price weighted, and the comparison is not a matched-budget intervention proving that one allocation is causally superior. Inherit-MAS is also not cheaper than Single ReAct because it performs task-specific search. 

\subsection{Detailed Cost Ledgers}
\label{sec:cost-details}
The tables in this subsection give the per-run provider ledgers, the inherited token equivalents, wall time, and the candidate-prefix costs.
\begin{table}[t]
\centering
\caption{Live provider usage on the full evaluation sets. Token values are totals in millions, and inherited token equivalents are excluded. Relative provider cost normalizes each benchmark to Single ReAct $=1.00\times$ using the configured deployment tariffs. On HotpotQA the Single ReAct harness also scores its one answer with the GPT-5.4-mini judge. Those calls do not change its output, and their cost is included in the reference, which raises the denominator of the HotpotQA relative-cost column. }
\label{tab:inference-cost}
\small
\setlength{\tabcolsep}{9pt}
\begin{tabular}{@{}lrrrr@{}}
\toprule
System & Worker tokens & Meta/judge tokens & Total tokens & Relative cost\\
\midrule
\multicolumn{5}{c}{\textit{WorkBench}}\\
\cmidrule(lr){1-5}
Single ReAct (reference) & 3.26 & 0 & 3.26 & $1.00\times$\\
\paddedhdashline
EvoAgent & 7.69 & 0.15 & 7.85 & $2.41\times$\\
EvoMAS & 168.48 & 5.11 & 173.59 & $70.15\times$\\
TacoMAS & 222.41 & 14.18 & 236.59 & $98.86\times$\\
\paddedhdashline
Inherit-MAS & 1.44 & 8.92 & 10.35 & $15.72\times$\\
\midrule
\multicolumn{5}{c}{\textit{HotpotQA}}\\
\cmidrule(lr){1-5}
Single ReAct (reference) & 1.35 & 0.23 & 1.58 & $1.00\times$\\
\paddedhdashline
EvoAgent & 3.68 & 0.66 & 4.34 & $1.24\times$\\
EvoMAS & 103.81 & 8.47 & 112.28 & $46.71\times$\\
TacoMAS & 9.00 & 8.37 & 17.37 & $22.08\times$\\
\paddedhdashline
Inherit-MAS & 3.60 & 4.59 & 8.18 & $9.86\times$\\
\bottomrule
\end{tabular}
\end{table}

\begin{table}[h]
\centering
\caption{Full-manifest inference and search cost. Total live includes every provider token. Inherited equivalents are not added. }
\label{tab:cost}
\scriptsize
\setlength{\tabcolsep}{3pt}
\begin{tabular}{llrrr}
\toprule
Benchmark & System & Total live & Inherited equiv. & Avg. live/task\\
\midrule
WorkBench & Single ReAct & 3.26M & 0 & 25.1k\\
WorkBench & EvoAgent & 7.85M & 0 & 60.3k\\
WorkBench & EvoMAS & 173.59M & 0 & 1.335M\\
WorkBench & TacoMAS & 236.59M & 0 & 1.820M\\
WorkBench & Inherit-MAS & 10.35M & 0.53M & 79.6k\\
\midrule
HotpotQA & Single ReAct & 1.58M & 0 & 5.3k\\
HotpotQA & EvoAgent & 4.34M & 0 & 14.5k\\
HotpotQA & EvoMAS & 112.28M & 0 & 374.3k\\
HotpotQA & TacoMAS & 17.37M & 0 & 57.9k\\
HotpotQA & Inherit-MAS & 8.18M & 1.93M & 27.3k\\
\bottomrule
\end{tabular}
\end{table}

\begin{table}[h]
\centering
\caption{Separated Inherit-MAS accounting. Wall is summed task wall. Inherited equivalents are excluded from live columns.}
\label{tab:ours-cost-ledger}
\scriptsize
\begin{tabular}{lrrrrr}
\toprule
Bench. & Worker & Meta/judge & Total & USD & Wall\\
\midrule
WorkBench & 1.44M & 8.92M & 10.35M & 8.07 & 1.41h\\
HotpotQA & 3.60M & 4.59M & 8.18M & 6.28 & 3.34h\\
\bottomrule
\end{tabular}
\end{table}

\subsection{Inheritance Ledger and Full Traversal}
\label{sec:ledger}
This section completes the accounting of \Cref{sec:inheritance}, using the inheritance indicator $z_{k,v}$ from \Cref{eq:hit}. The two structural cases of change localization have a direct interpretation. For $v\notin D_k$, the runtime verifies that the current request digest agrees with the incumbent node record before using its result. For $v\in D_k$, it constructs the request from the current upstream results and queries the same result store. An affected node can therefore still produce an inheritance hit when an affected path resolves to a request seen earlier. Otherwise, the node executes live and its eligible result is inserted as a new immutable record. An unchanged structure identifies the expected incumbent record, while the complete resolved-request digest remains the sole authority for execution inheritance.

If $t_{k,v}$ is the original worker-token usage associated with the stored or live result, the two ledgers are
\begin{equation}
\begin{aligned}
T_{\mathrm{live}}&=\sum_{k,v}(1-z_{k,v})t_{k,v},\\
T_{\mathrm{inherited}}&=\sum_{k,v}z_{k,v}t_{k,v},\\
\rho&=\frac{T_{\mathrm{live}}}
{T_{\mathrm{live}}+T_{\mathrm{inherited}}}.
\end{aligned}
\label{eq:ledger}
\end{equation}
Here $\rho$ is the fraction of the trajectory's worker-token volume executed live. An inheritance hit contributes zero live worker tokens. $T_{\mathrm{inherited}}$ is a counterfactual token-equivalent and is never added to provider usage. A hit uses the previously observed result rather than drawing a new sample; the Inherit-MAS experiments therefore use temperature zero. Provider-side prompt caching is complementary because it still invokes the model, whereas execution inheritance removes the call.

Full execution executes every node. Full traversal with the same result store resolves every request digest before lookup. Inherit-MAS additionally uses the affected region to distinguish structurally unchanged nodes from request reconvergence inside that region, while still verifying a complete digest in both cases. Full traversal and the change-localized schedule therefore have the same live model-call miss set $\{(k,v):z_{k,v}=0\}$. We attribute worker-token savings to verified execution inheritance; the affected region provides change localization, auditability, and subgraph-level accounting rather than a separate source of worker-token savings.

\subsection{Full-Execution Reruns and Store Footprint}
\label{sec:store-footprint}
The full-execution bars in \Cref{fig:full-eval-cost} come from independent reruns of the same Inherit-MAS controller with execution inheritance disabled, so every node executes live and the ledgers record actual provider usage. The two WorkBench reruns use 2.08M and 1.98M worker tokens and the HotpotQA rerun 5.50M, against 1.44M and 3.60M for the original runs. The reruns also check the inherited token equivalents recorded by the original runs. Charging every inheritance hit its stored original usage predicts 1.97M and 5.53M worker tokens under full execution, within 3\% of the mean WorkBench rerun usage and the HotpotQA rerun usage, respectively. The deterministic-runtime audit below provides the evidence for execution equivalence under its tested conditions.

The persisted task-local result stores contain 979 WorkBench entries occupying 5.14 MiB (5.51 kB/entry) and 2,694 HotpotQA entries occupying 17.40 MiB (6.77 kB/entry). A post-hoc warm local-filesystem microbenchmark of
record read, JSON decode, SHA-256, and in-memory rehydration takes median
0.080 ms (p95 0.122 ms) per WorkBench entry and 0.062 ms (p95 0.344 ms) per
HotpotQA entry. These timings report storage-system scale only. They are not
online latency measurements or evidence that the change-localized schedule outperforms full
traversal with the same result store.

\paragraph{Deterministic inheritance qualification.}
Before the remote-API runs, we qualified the three execution conditions (full execution, full traversal with the result store, and the change-localized schedule) on a
local deterministic Qwen3-14B and LiveCodeBench substrate. Across 240 frozen
candidate-problem pairs, full execution, full traversal with the result store,
and the change-localized schedule produced byte-identical node outputs and score vectors.
The two inheritance conditions also had identical model-call sets and each used
17.7\% of the full-execution worker-token volume. This audit
validates request-verified execution inheritance on a qualified
deterministic runtime. For the remote models, inheritance remains scoped to
the stored request and output and does not assert fresh-call byte identity.

\subsection{Candidate-Prefix Cost}
\Cref{tab:prefix-cost} reports cumulative live cost as the test-time trajectory
grows. Model wall sums worker execution and meta-model and judge call
latencies and excludes deterministic orchestration. The fifth candidate is
more expensive because the final-round restart frequently produces a new workflow
rather than applying an inheritance-friendly local edit.

\begin{table}[h]
\centering
\caption{Cumulative live cost through candidate prefixes. Tok. is total
provider tokens. Model wall is summed model-call wall time.}
\label{tab:prefix-cost}
\scriptsize
\begin{tabular}{llrrr}
\toprule
Benchmark & Prefix & Tok. & USD & Model wall\\
\midrule
WorkBench & 1 & 2.20M & 1.77 & 0.38h\\
 & 3 & 6.24M & 4.87 & 0.86h\\
 & 5 & 10.35M & 8.07 & 1.39h\\
\midrule
HotpotQA & 1 & 1.84M & 1.22 & 0.86h\\
 & 3 & 4.97M & 3.72 & 1.98h\\
 & 5 & 8.18M & 6.28 & 3.30h\\
\bottomrule
\end{tabular}
\end{table}

\subsection{Artifacts and Accounting}
Every run records a frozen manifest digest, source/configuration fingerprints, per-call provider usage, task-level wall time, tool or retrieval calls, invalid attempts, and atomic completion records. Candidate-generation, judge, worker, deterministic-tool, and benchmark-evaluator ledgers remain separate. Provider cache counters are recorded but are not treated as Inherit-MAS execution inheritance. A request-verified hit carries the original node usage as \inherited{} and records zero \live{}. Stored record content and identity are digest-checked on every load. Any mismatch is a blocking integrity error.

\section{Code and Data Availability}
The implementation, evaluation task lists, and scoring scripts are available at \url{https://github.com/CrazyMint/Inherit-MAS}. WorkBench and HotpotQA data remain subject to their original licenses; restricted source data will not be redistributed.

\end{document}